\documentclass[journal,transmag]{IEEEtran}

\ifCLASSINFOpdf
\else
\fi

\usepackage{cite}
\usepackage{graphicx}
\usepackage{bm}
\usepackage{multirow}
\usepackage{booktabs}
\usepackage{amssymb}

\usepackage{algorithm} 
\usepackage{algorithmic}
\usepackage{bm}

\usepackage{subfigure}
 \usepackage{amsmath} % Gabor added, to make multi-row equation more elegant

\begin{document}
	\title{3DMNDT: 3D multi-view registration method based on the normal distributions transform}

	
	\author{\IEEEauthorblockN{Jihua Zhu\IEEEauthorrefmark{*1},
			Di Wang\IEEEauthorrefmark{1}, Jiaxi Mu\IEEEauthorrefmark{2}, Huimin Lu\IEEEauthorrefmark{3}, Zhiqiang Tian\IEEEauthorrefmark{1}, Zhongyu Li\IEEEauthorrefmark{1}, }
		\IEEEauthorblockA{\IEEEauthorrefmark{1} School of Software Engineering, Xi'an Jiaotong University, China\\
		\IEEEauthorrefmark{2} School of Artificial Intelligence, Xi'an Jiaotong University, China\\
		\IEEEauthorrefmark{2} Kyushu Institute of Technology, Japan}% <-this % stops an unwanted space
		\thanks{* Corresponding author: Jihua Zhu (email: zhujh@xjtu.edu.cn).}}

	\markboth{Journal of \LaTeX\ Class Files,~Vol.~14, No.~8, March~2021}%
	{Shell \MakeLowercase{\textit{et al.}}: Bare Demo of IEEEtran.cls for IEEE Transactions on Magnetics Journals}



	% for Transactions on Magnetics papers, we must declare the abstract and
	% index terms PRIOR to the title within the \IEEEtitleabstractindextext
	% IEEEtran command as these need to go into the title area created by
	\IEEEtitleabstractindextext{%
		\begin{abstract}
       The normal distributions transform (NDT) is an effective paradigm for the point set registration. This method is originally designed for pair-wise registration and it will suffer from great challenges when applied to multi-view registration. Under the NDT framework, this paper proposes a novel multi-view registration method, named 3D multi-view registration based on the normal distributions transform (3DMNDT), which integrates the K-means clustering and Lie algebra solver to achieve multi-view registration. More specifically, the multi-view registration is first cast into the problem of maximum likelihood estimation. Then, the K-means algorithm is utilized to divide all data points into different clusters, where a normal distribution is computed to locally models the probability of measuring a data point in each cluster. Subsequently, the registration problem is formulated by the NDT-based likelihood function. To maximize this likelihood function, the Lie algebra solver is developed to sequentially optimize each rigid transformation. The proposed method alternately implements data point clustering, NDT computing, and likelihood maximization until desired registration results are obtained. Experimental results tested on benchmark data sets illustrate that the proposed method can achieve state-of-the-art performance for multi-view registration.
		\end{abstract}
		
		}

	\maketitle

	\IEEEdisplaynontitleabstractindextext
	
	\IEEEpeerreviewmaketitle

	\section{Introduction}
\IEEEPARstart{P}{oint} set registration is a fundamental problem arising from main domains, such as computer vision \cite{sandhu2009point,yang2015go,ma2019locality}, robotics \cite{jiang2019simultaneous,tabib2018manifold,jiang20203d}, and medical image analysis \cite{rasoulian2012group,ravikumar2018group}. With the development of scanning technology, many scan devices are able to acquire accurate 3D scan data from a scene or object. However, they are difficult to scan the whole object or scene at one time due to the limited field of view. For 3D reconstruction, it is necessary to acquire scan data at different viewpoints and align these scans, which arises the problem of point set registration. Given the reference frame, the goal of multi-view registration is to compute the spatial transformations of each point set, so as to unify them into the same coordinate system. 
	
In the literature, Iterative Closest Point (ICP) \cite{besl1992method} is one of the most popular solutions to point set registration. Given initial rigid transformation, this method alternatively establishes point correspondence and  optimizes rigid transformation to achieve pair-wise registration with good accuracy. However, the original ICP method is unable to register partially overlapping point sets and it is time-consuming due to the registration over raw data points. Then, the normal distributions transform (NDT) \cite{biber2003normal,magnusson2009three} is proposed to solve pair-wise registration problem. For the pair-wise registration, it divides the space occupied by the target point set into regularized cells and approximates data points in each cell by the normal distribution. Further, each data point of source point set is assigned to one NDT and then the registration problem is formulated by the NDT-based objective function, which is optimized by the Newton’s 
solver. Since one NDT is utilized to approximate all data points in each cell, this method is efficient to align large scale data sets. What's more, it is able to align partially overlapping point sets. Therefore, it is very popular in computer vision. 
	
Although NDT has been widely applied to pair-wise registration, it has not been extended to 3D multi-view registration, which suffers from great challenges. Firstly, it is difficult to directly formulate the multi-view registration problem and simultaneously optimize a lot of rigid transformations under the original NDT framework. Secondly, the original NDT divides the space into regularized cells. This regularized division leads to the large variance of the number of points in each cell, which reduces the registration performance of NDT. Thirdly, the original NDT utilizes the Newton’s solver, which is nasty to directly compute Jacobian matrix and Hessian matrix for the optimization of 3D rigid transformations. 

In this paper, we extend the original NDT and propose a novel method named 3DMNDT to solve the 3D multi-view registration problem. The contributions of this paper are delivered as follows: 1) We formulate the multi-view registration problem into NDT based likelihood function. Therefore, multi-view registration can be achieved by the maximum likelihood estimation. 2) We utilize the K-means clustering to replace the regularized division of space cubes, which can avoid the unbalanced division of data points. 3) We propose the Lie algebra solver to sequentially optimize each rigid transformation. Compared with the Newton’s 
solver, this optimization algorithm avoids the calculation of Jacobian matrix and Hessian matrix. 4) We test 3DMNDT on seven benchmark data sets and compare it with several state-of-the-art methods. Experimental results illustrate its superior performance for 3D multi-view registration. 

The remainder of this paper is organized as follows. Section 2 briefly discusses related methods of point set registration. In Section 3, we formulate the multi-view registration problem under the NDT framework and then derive the 3DMNDT to solve 3D multi-view registration. We also analyze its computation complexity and present its implementation details. Section 5 displays the experimental results on benchmark data sets. Finally, we conclude this paper in Section 6.

\section{Related work}
This section only discusses these methods related to the
proposed method for multi-view registration. For convenience, the terms motion and rigid transformation are utilized interchangeably throughout this paper.

According to the number of point sets being registered, point set registration can be divided into pair-wise registration problem and multi-view registration problem. For pair-wise registration, many variants have been derived from the original ICP algorithm and they improve the performance of ICP from different aspects \cite{rusinkiewicz2001efficient}. For partially overlapping point sets, the Trimmed ICP (TrICP) algorithm \cite{chetverikov2005robust} introduces the overlap parameter to automatically distinguish overlapping regions and is able to achieve good registration. For accurate registration, Myronenko and Song proposed the Coherent Point Drift (CPD) algorithm \cite{myronenko2010point}, which considers pair-wise registration as a probability density estimation problem. More specifically, it takes the source point set as the Gaussian mixture model (GMM) centroids and fits them to the target point set by maximizing the likelihood. Besides, Jian et al. \cite{jian2010robust} formulated the pair-wise registration problem as aligning two GMMs, where each point set is represented by one GMM. Then it achieves pair-wise registration by minimizing $L_2$ measure between two GMMs. Similar to CPD, Gao and Tedrake \cite{gao2019filterreg} proposed the FilterReg, which takes the target point set as GMM centroids and fits the source point set to GMM by maximizing the likelihood. Although these GMM-based methods are powerful for both rigid and non-rigid registration, it requires to build all point correspondences between two point sets being registered, which leads to high computational complexity.

Although these methods are effective for pair-wise registration, they are local convergent and can only obtain desired registration results in the case of provided with good initial parameters. To estimate initial parameters, Sipiran and Bustos \cite{sipiran2011harris} proposed the 3D Harris detection method, which can detect the interest-point from 3D point set represented by meshes. Given detected interest points, feature descriptors with match method \cite{rusu2009fast,guo2013rotational,lei2017fast} can be applied to achieve pair-wise registration for point sets with arbitrary orientations. Recently, some deep learning methods \cite{aoki2019pointnetlk,wang2019deep,choy2020deep} have been proposed to achieve global pair-wise registration. Although some of these methods are effective, most of them are supervised learning method, which requires many data sets to train the registration model. What's more, these methods cannot be directly extended to solve multi-view registration problem. 

Usually, multi-view registration is much more complex than pair-wise registration but receives less attention. For multi-view registration, the primary method \cite{chen1992object} is to incrementally register and merge point set pairs until all point sets are merged into one model. Although this method is easy to operate, it suffers from the accumulation error problem with the increase of merged point sets. Then, Bergevin et al.~\cite{bergevin1996towards} proposed the first method for multi-view registration. This method sequentially optimizes the rigid transformation over the established point correspondences, which are built between one point set between each other point set. This method overcomes the problem of cumulative error, but it neglects non-overlapping regions between each point set pairs, which reduces the registration accuracy. To address this issue, Zhu et al. \cite{zhu2014surface} proposed the coarse-to-fine registration method, which sequentially optimizes the trimmed errors of point correspondences, which built between one point set and other aligned point sets. With the increase of point sets, the coarse-to-fine registration method tends to be trapped into local minimum. Accordingly, Tang et al. \cite{tang2015hierarchical} proposed hierarchical multi-view registration method, which continuously implements multi-view registration on small number of point sets and merges them as the one in the next implementation. 

As there are many rigid transformations involved in multi-view registration, most of methods utilize alternating optimization strategy to sequentially optimize each rigid transformation. However, some methods utilize the batch optimization strategy. Given point correspondences established between each point set pair,
Krishnan et al. \cite{krishnan2005global} proposed a multi-view registration method via optimization-on-a-manifold, which can simultaneously optimize all rigid transformations. However, it is difficult to obtain point correspondences between each point set pair, which should be provided in advance for this method. Accordingly, Mateo et al. \cite{mateo2014bayesian} treated pairwise correspondences as missing data and formulated multi-view registration as the maximum likelihood estimation, where all rigid transformations are optimized by the Expectation–Maximization (EM) algorithm in batch mode. Since there are massive point correspondences, this method requires to estimate massive hidden variables, which is time-consuming. 

Since multi-view registration is more difficult than pair-wise registration, one feasible method is to recover the global motions from relative motions, where the term motion denotes rigid transformation. Accordingly, Govindu et al. \cite{govindu2013averaging} proposed the motion averaging algorithm and applied it to multi-view registration. This method utilizes the Lie group structure of motions \cite{sola2018micro} to implement the averaging of all available relative motions. Theoretically, accurate global motions can be obtained from the number of relative motions, which is equal to or more than the number of point sets being registered. However, this method is sensitive to outliers, where even one outlier will lead to the failure of multi-view registration. Further, Arrigoni et al. \cite{arrigoni2016global, arrigoni2018robust} cast multi-view registration to a low-rank and sparse (LRS) matrix decomposition problem, where registration results are recovered from the low-rank matrix. In this method, it requires to concatenate all available relative motions as well as some zero matrices into one large matrix, which is then decomposed into a low-rank matrix and a sparse matrix. This method is robust to outliers but it tends to be failed when the concatenated matrix is sparse \cite{arrigoni2018robust}. Besides, these two methods equally treat all relative motions, which may possess different reliabilities. Accordingly, Guo et al. \cite{guo2018weighted} proposed weighed motion averaging algorithm and Jin et al. \cite{jin2018multi} proposed the weighted LRS for multi-view registration. Since these two methods pay more attention to these relative motions with high reliability, they are more likely to obtain desired registration results than their original methods.

Recently, Georgios et al. \cite{evangelidis2017joint} cast multi-view registration to the clustering problem and proposed JRMPC to solve this special problem. This method assumes that all data points are generated from a central GMM and utilizes the EM algorithm to estimate the GMM components as well as rigid transformations. Since this method requires to estimate massive model parameters related to all data points, it is time-consuming and likely to be trapped into local minimum. To address these issues, Zhu et al. \cite{zhu2019efficient} extended the k-means algorithm to solve multi-view registration problem. Compared with JRMPC, this method is efficient and more likely to obtain desired registration results. However, it utilizes one cluster centroid to approximate all data points in the same cluster, which inevitably leads to a lot of information loss and reduces the registration accuracy. Further, Zhu et al. \cite{zhu2020registration} proposed an efficient GMM-based method under the perspective of EM. This method assumes that each data point is drawn from its corresponding GMM, where its nearest neighbors ( NNs ) in other point sets are taken as Gaussian centroids with similar covariance and membership probability. Since this method only requires to estimate quite a few rigid transformations as well as one covariance, it is efficient than other mixture model-based methods. However, these mixture model-based methods are more time-consuming than most previous methods.

The original NDT was first proposed to solve the registration of 2D point sets in robotics \cite{biber2003normal}. As this method utilizes the ND to approximate a set of point sets, it is efficient to achieve pair-wise registration. Further, it was extended to align 3D point sets \cite{magnusson2009three}, where the computation of Jacobian and Hessian matrix with respect to Euler angles is really nasty. Besides, the spatial division mechanism leads to regular cells, which reduces the registration performance. Accordingly, Arun and Steven \cite{das2012scan} replaced regular division by k-means clustering to divide all data points of target point set into different clusters, where the target point set is always static during registration. However, this method cannot be directly extended to multi-view registration, where rigid transformations of point set are constantly changed and all data points should be dynamically clustered at each iteration. In this paper, we formulate multi-view registration problem into the NDT-based objective function, which is then maximized by the integration of the k-means clustering and the Lie algebra solver.    

\section{Problem formulation}

Let ${\bf V}_i = [{{\bf{v}}_{i,1}}...{{\bf{v}}_{i,j}}...{{\bf{v}}_{i,{N_i}}}]$ indicates $N_i$ data points in the $i$-th point set and ${\mathbb V} = \{ {{\bf V}_i}\} _{i = 1}^M$ denotes $M$ point sets being registered. Similar to some previous methods, we regard multi-view registration as the clustering problem. Given the reference frame, the goal of multi-view registration is to simultaneously divide all data points into $K$ clusters and estimate accurate rigid transformations ${\mathbb T}= \{ {\bf R}_i \in {\rm SO(3)} ,{\bf t}_i \in {\mathbb R}^3 \}_{i = 1}^M$ for each point set. For convenience, the rigid transformation is concatenated into motion as follows:
\begin{equation}
      {\cal T}_i = 
      \begin{bmatrix}
      {\bf{R}}_i&{\bf{t}}_i\\
    {\bf 0} &1
      \end{bmatrix}, 
\end{equation}
where $ {\cal T}_i \in {\rm SE(3)}$. Besides, we define the group action of ${\rm SE(3)}$ on a 3D vector ${\bf v}_{i,j}$ as ${\cal T}_i {\bf v}_{i,j} = {\bf R}_i {\bf v}_{i,j} + {\bf t}_i$. For each cluster, it is convenient to compute the NDT $({\bm \mu}_k, {\bm \Omega}_k)$, which denote the mean (centroid) and information matrix (the inverse of covariance matrix), respectively. 

Suppose the aligned data point ${{\cal T}_i}{{\bf{v}}_{i,j}}$ belongs the $c_{(i,j)}$-th cluster, then it is reasonable to assume that the aligned data point ${{\cal T}_i}{{\bf{v}}_{i,j}}$ is drawn from the normal distribution $ {\cal T}_i {\bf v}_{i, j} \sim  {\cal N} ( {\bm \mu}_{c(i, j)}, {\bm \Omega}_{c(i, j)}^{-1} ) $, where $c_{(i,j)}$ denotes the index of cluster, which contains the data point ${{\bf{v}}_{i,j}}$. 
Therefore, the probability function is defined for ${{\bf{v}}_{i,j}}$ as follows:
\begin{equation}
p({\bf v}_{i, j}) = {\cal N} ( {\cal T}_i {\bf v}_{i, j} ; {\bm \mu}_{c(i, j)}, {\bm \Omega}^{-1}_{c(i, j)}). 
\end{equation}
Further, all data points are assumed to be independent of each other and their joint probability distribution is formulated as:
\begin{equation}
p({\mathbb V}) = \prod_{i=1}^M  \prod_{j=1}^{N_i}  {\cal N}( {\cal T}_i {\bf v}_{i, j} ; {\bm \mu}_{c(i, j)}, {\bm \Omega}_{c(i, j)}^{-1}),
\end{equation}
where the normal distribution on ${\mathbb R}^3$ is defined as:
\begin{equation}\nonumber
{\cal N}( {\bf x} ; {\bm \mu}, {\bm \Omega}^{-1} ) = \sqrt{ \frac{ |{\bm \Omega}|  }{ (2\pi)^3 } } \exp \{  -\frac{1}{2} ({\bf x} - {\bm \mu} )^\top {\bm \Omega} ({\bf x} - {\bm \mu} ) \}.
\end{equation}

This joint probability distribution is utilized to define the logarithm of likelihood function as:
\begin{equation}
\begin{aligned}
L({\bm \Theta}) &= \log p({\mathbb V}) \\
&= \sum_{i = 1}^M \sum_{j = 1}^{N_i } \log  {\cal N} ( {\cal T}_i {\bf v}_{c(i, j)} ; {\bm \mu}_{c(i, j)}, {\bm \Omega}_{c(i, j)}^{-1} ),
\end{aligned}
\label{eq:LLH}
\end{equation}
where ${\bm \Theta}  \buildrel \Delta \over = \left\{ 
{\left( {\cal T}_i \right)_{i = 1}^M,} {\left( {{{\bm{\mu }}_k},{{\bm \Omega}_k}} \right)_{k = 1}^K}\right\}$ denotes the model parameters.

Accordingly, the multi-view registration is transformed into the maximum likelihood estimation, which is formulated as:
\begin{equation}
\begin{aligned}
&\max_{\bm \Theta} L({\bm \Theta})\\
&{\rm s.t. } \quad {\cal T}_i \in {\rm SE(3)}, i = 1, 2, \cdots, M.
\end{aligned}
\label{eq:obj}
\end{equation}
% \quad  {\bf R}_i^\top {\bf R}_i = {\bf I}_3, {\rm det}({\bf R}_i)  = 1
By the optimization of Eq. (\ref{eq:obj}), multi-view registration can be achieved with data point clustering.

\begin{figure*}[!t]
		\centering
		\includegraphics[width=1.0\textwidth]{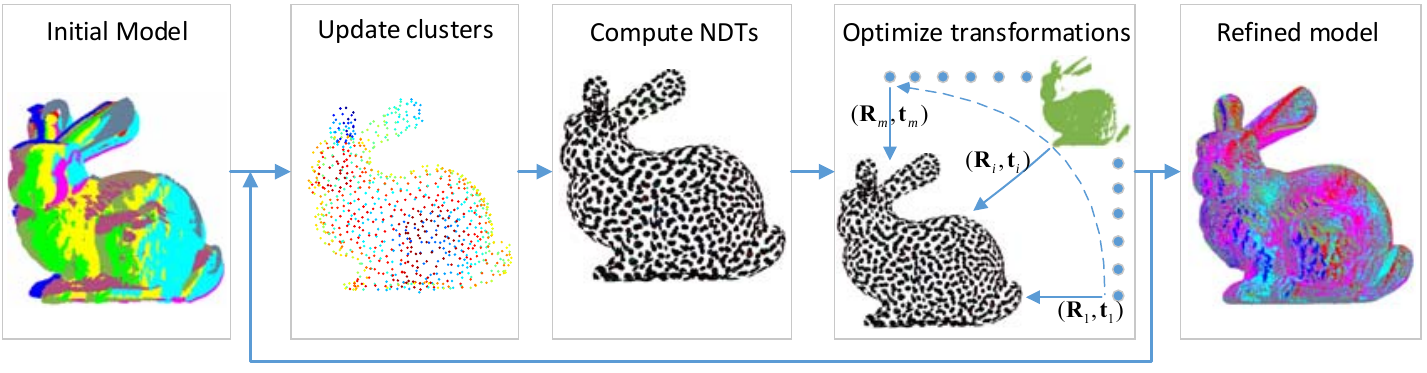}
		\caption{Illustration of the proposed method, which integrates the K-means clustering and Lie algebra solver to achieve multi-view registration under the NDT framework. Given initial model parameters, the proposed method uses the K-means clustering algorithm to divide all aligned data points into different clusters and compute the NDT for each cluster. Then it utilizes the Lie algebra solver to sequentially optimize each rigid transformation. The desired registration results will be obtained by alternatively implementing data point clustering, NDT computation, and rigid transformation optimization.}
		\label{fig:3DMNDT}
\end{figure*}

\section{3DMNDT method}

For the optimization of Eq. (\ref{eq:obj}), we propose the 3D multi-view version of NDT method (3DMNDT) displayed in Fig. \ref{fig:3DMNDT}. As shown in Fig. \ref{fig:3DMNDT}, this method optimizes Eq. (\ref{eq:obj}) by iterations. Given initial rigid transformations $ ( {\cal T}^0_i )_{i = 1}^M$ and initial cluster centroids $( {\bm \mu}_k^0 )_{k = 1}^K$ , three steps are alternately executed in each iteration. 

(1)	Divide all data points into different clusters.
\begin{equation}
     c^h(i,j) = \mathop {\arg \min }_{k  = 1,2, \cdots, K } {\left\| {\cal T}_i^{h-1} {\bf v}_{i, j} - {\bm \mu}_k^{h - 1} \right\|_2^2}.
     \label{eq:corr1}
\end{equation}
    
(2)	Compute the NDT for each cluster.
\begin{equation}
{\bm \mu}_k^h = \frac{ \sum_{i = 1}^M  \sum_{j = 1}^{N_i} \{ c^h(i, j) = k \} {\cal T}_i^{h-1} {\bf v}_{i, j} } { \sum_{i = 1}^M \sum_{j = 1}^{N_i} \{ c^h(i,j) = k \} },
\label{eq:mu}
\end{equation}
\begin{equation}
{\bm \Omega}_k^h = ( {\bm \Sigma} _k^h + \varepsilon {\bf I}_3 )^{ - 1},
\label{eq:inf}
\end{equation}
where $\varepsilon  = 10^{ - 6}$ prevents the occurrence of zero denominator and ${\bm \Sigma}^h_k$ indicates the covariance matrix calculated by:
\begin{equation}
{\bm \Sigma}_k^h = \frac{ \sum_{i = 1}^M  \sum_{j = 1}^{ N_i } \{ c^h (i,j) = k \}  \Delta {\bf v}_{i,j}^h   (\Delta{\bf v}_{i,j}^h )^\top} { \sum_{i = 1}^M \sum_{j = 1}^{ N_i } \{  c^h(i,j) = k \} },
\end{equation}
where $\Delta {\bf v}_{i,j}^h =  {\cal T}_i^{h - 1} {\bf v}_{i, j} - {\bm \mu}_k^h$. 

It should be noted that some clusters may only contain a small number of data points. In 3D cases, points of one cluster may be perfectly co-planar, the covariance matrix will always be singular and can not be inverted. For this reason, NDTs only computed for clusters containing more than 5 points. Meanwhile, other clusters are regarded as invalid clusters. 

(3) Rigid transformation optimization.

If one data point belongs to a invalid cluster, it should be neglected in the optimization of rigid transformation. Accordingly, all rigid transformations can be optimized by maximum likelihood estimation: 
\begin{equation}
 \max_{ \mathbb T} \sum\limits_{i = 1}^M \sum_{j = 1}^{N_i^{'}}  \log {\cal N}({\cal T}_i {\bf v}_{i, j}; {\bm \mu }_{c^h(i,j)},{\bm \Omega}^{-1}_{c^h(i,j)} ),
\end{equation}
where $N^{'}_i\le N_i$ denotes the number of valid data point in the $i$-th point set. 

To obtain the desired results, the proposed method should iteratively operate these three steps until the likelihood $L({\bm \Theta})$ has no significant change or the iteration number $h$ reaches the maximum value. Obviously, Steps 1 denotes the nearest neighbor search problem, which can be efficiently solved by the $k$-d tree based method \cite{nuchter2007cached}. Step 2 is a trivial calculation problem. However, Step 3 remains to be a difficult optimization problem, which is the most critical step of the proposed method. 

\subsection{Optimizing by the Lie algebra solver}
It seems difficult to simultaneously optimize all rigid transformations involved in Eq. (\ref{eq:obj}). Given clustering results, each rigid transformation can be separately optimized by the following objective function:
\begin{equation}
\max_{ {\cal T}_i \in {\rm SE(3)}} \sum_{j = 1}^{N_i^{'} }  \log {\cal N} ({\cal T}_i{\bf v}_{i,j} ; {\bm \mu}_{c^h(i,j)}, {\bm \Omega}^{-1}_{c^h(i,j)} ), 
\end{equation}
where $i=1, 2, \cdots, M$.
For simplicity, we discard the iteration number $h$, the point set number $i$ and cluster number $c(i, j)$ in all variables. Accordingly, the above optimization can be simplified as a non-linear least-squares problem: 
\begin{equation}
\begin{aligned}
\max_{ {\cal T} \in {\rm SE(3)} }  L({\cal T}) &= \max_{ {\cal T} \in {\rm SE(3)} } \sum_{j = 1}^N \log {\cal N} ( {\cal T} {\bf v}_j;  {\bm \mu}_j, {\bm \Omega}^{-1}_j )  \\
&= \min_{{\cal T} \in {\rm SE(3)}} \sum_{j = 1}^N  {\bf r}_j ^\top{\bm \Omega }_j {\bf r}_j , 
\end{aligned}
\label{eq:objf}
\end{equation}
where ${\bf r}_{j} = {\cal T} {\bf v}_j  - {\bm \mu}_j$ indicates the residual error between the data point ${\cal T}{\bf v}_{j}$ and its cluster centroid ${\bm \mu}_{j}$. 

 To circumvent the constraints stems from rigid transformation ${\cal T}$, one feasible method is to use Lie algebra as follows:
\begin{equation}
\max_{{\cal T} \in {\rm SE(3)}} L({\cal T}) = \max_{ {\bm \xi} \in {\mathbb R}^6 } L( \exp ({\bm \xi}^\wedge) {\cal T}^0), 
\end{equation}
where ${\cal T}^0$ denotes rigid transformation estimated from previous iteration, ${\bm \xi}$ denotes the vector space of Lie algebra ${\mathfrak se(3)}$,  $(\cdot)^\wedge : {\mathbb R}^6 \rightarrow {\mathbb R}^{4 \times 4}$ denotes the skew operator which converts vector space ${\mathbb R}^6$ to Lie algebra ${\mathfrak se(3)}$, $\exp(\cdot): {\mathfrak se(3)}  \rightarrow {\rm SE(3)}$ denotes the exponential map which lifts Lie algebra to Lie Group. As the exponential map of ${\mathfrak se(3)}$ involves the calculation of Jacobian matrix, it is quite complex and highly non-linear. Accordingly, we turn to utilize retraction technique \cite{forster2016manifold}.

%The most important property of Lie group is that it has Lie algebra $\mathfrak{se}(3)$, which is the tangent space at the identity of ${\rm SE(3)}$. The Lie algebra $\mathfrak{se}(3)$ is a vector space and it can be compactly denoted as a vector ${\bm \xi} \in {\mathbb R}^6$. By utilizing Lie algebra instead of Lie group, we can easily to avoid the constraint stems of rigid transformation and cast the original constrained optimization to unconstrained optimization with respect to Euclidean space. Subsequently, the optimized Lie algebra can be converted into Lie Group using the exponential map $\exp(\cdot): {\mathbb R}^6 \rightarrow {\rm SE(3)}$. 
%To optimize the objective function, the original NDT requires to utilize the exponential map of ${\rm SE(3)}$, which is quite complex and highly non-linear due to the Jacobian matrix. 

% For the convenience of optimization, we turn to utilize retraction technique \cite{forster2016manifold}. 

In the retraction technique, the vector space of ${\mathfrak se(3)}$ is decomposed into rotational and translational components, i.e., ${\bm \xi} = \{ {\bm \xi}_R \in {\mathbb R}^3, {\bm \xi}_t \in {\mathbb R}^3 \}$, and then the retraction technique ${\cal R}(\cdot): {\mathbb R}^6 \rightarrow {\rm SE(3)} $ lifts ${\bm \xi}$ to ${\rm SE(3)}$ as follows:
\begin{equation}
{\cal R}({\bm \xi}) = 
\begin{bmatrix}
\exp({\bm \xi}_R^\wedge)   &{\bm \xi}_t \\
{\bf 0}         &1
\end{bmatrix},
\label{eq:RetractionOp}
\end{equation}
where the exponential map $\exp( \cdot ): {\rm so(3)} \rightarrow {\rm SO(3)}$ lifts ${\mathfrak so}(3)$ to ${\rm SO(3)}$. It is worth noting that the skew operator has an important property: 
\begin{equation}
{\bm \xi} _R^ \wedge {\bf v} =  - {\bf v}^\wedge {\bm \xi}_R, \forall {\bf v} \in {\mathbb R}^3, 
\label{eq:pro}
\end{equation}
which will be useful later on.
 
 % which can be mapped to the Lie group by exponential operator $\exp ( \cdot )$
% and coincides with the space of $4 \times 4$ skew symmetric matrix. 
% Given a vector ${\bm{\xi }}$, each skew symmetric matrix is identified by the hat operator $\wedge$ as follows:
% \begin{equation}
%{{\bf{\xi }}^{\wedge}} = {\left[ {\begin{array}{*{20}{c}}
%{{\xi _{\bf{R}}}}\\
%{{\xi _{\bf{t}}}}
%\end{array}} \right]^{\wedge}} = \left[ {\begin{array}{*{20}{c}}
%0&{ - {\xi _{{\bf{R}},3}}}&{{\xi _{{\bf{R}},2}}}&{{\xi _{{\bf{t}},1}}}\\
%{{\xi _{{\bf{R}},1}}}&0&{ - {\xi _{{\bf{R}},1}}}&{{\xi _{{\bf{t}},2}}}\\
%{ - {\xi _{{\bf{R}},2}}}&{{\xi _{{\bf{R}},1}}}&0&{{\xi _{{\bf{t}},3}}}\\
%0&0&0&1
%\end{array}} \right]
%\end{equation}
%where ${\cal T}= exp({{\bf{\xi }}^{\wedge}})$, 
%$\mathfrak{se}(3) = \{ \mathbf{\xi } \in \mathbb{R}^{6} \} $ and $\mathfrak{so}(3) = \{ {\bm \xi}_R \in \mathbb{R}^{3}\}$ denotes the Lie algebra of $SE(3)$ and $SO(3)$, respectively.
%Conversely, the Lie group is mapped to the Lie algebra by the logarithmic operator $\log ( \cdot )$ and a skew symmetric matrix is mapped to a vector by the vee operator ${( \cdot )^ \vee }$, such as $\log {({\cal T})^ \vee } = \xi $. 

Since the group action of ${\rm SE(3)}$ is matrix multiplication, we denote the current rigid transformation as ${\cal T}^0$, and assume the perturbation of rigid transformation is ${\bm \xi} \in {\mathbb R}^6$. Consequently, the optimization is with respect to ${\bm \xi}$, and the rigid transformation is updated as follows: 
\begin{equation}
{\cal T}^* = {\cal R}({\bm \xi}^*) {\cal T}^0, 
\label{eq:pertu}
\end{equation}
where ${\bm \xi}^*$ denotes the optimized perturbation. 

Subsequently, we present the following theoretical result.
\textbf{Proposition 1} \textit{By utilizing retraction technique in Eq. (\ref{eq:RetractionOp}), the optimal perturbation can be attained by solving the following quadratic program (QP) problem:
\begin{equation}
\begin{aligned}
{\bm \xi}^* &= \mathop{\arg \min}\limits_{ {\bm \xi} \in {\mathbb R}^6 }  {\bm \xi}^\top {\bf H} {\bm \xi} + 2 {\bf  b}^\top {\bm \xi} + c\\
{\bf H} &= \sum_{j = 1}^N {\bf H}_j^\top  {\bm\Omega}_j  {\bf H }_j  \in  {\mathbb R}^{6 \times 6}\\
{\bf b} &= \sum_{j = 1}^N {\bf H}_j^\top {\bm \Omega}_{j}  {\bf  r}_j^0   \in {\mathbb R}^6 \\
c &= \sum_{j = 1}^N  ( {\bf r} _j^0) ^\top {\bm\Omega }_j {\bf r }_j^0  \in {\mathbb R}^ +
\end{aligned}
\label{eq:qp}
\end{equation}
where ${\bf H}_j \buildrel \Delta \over = \left[  -\left( {\bf R}^0 {\bf v}_j + {\bf t}^0 \right)^ \wedge  \quad {\bf I} \right] \in {\mathbb R}^{3 \times 6}$ and ${\bf r}_j^0 = {\cal T}^0 {\bf v}_j - {\bm \mu}_j$.}

\textbf{Proof 1} \textit{ Based on the assumption of small perturbation, the exponential operator of ${\mathfrak so}(3)$ can be approximated as:
\begin{equation}
\exp \left( {\bm \xi}_R^{\wedge} \right) \approx {\bf I} + {\bm \xi}_R^{\wedge}.
\label{eq:appr}
\end{equation}
We can substitute the Eq. (\ref{eq:appr}) into the Eq. (\ref{eq:pertu}), which leads to the rigid transformation defined as: 
\begin{equation}
\begin{aligned}
{\bf R} &=  \exp ({\bm \xi}_R^{\wedge} ){\bf R}^0 \approx {\bf R}^0 + {\bm \xi}_R^{\wedge}  {\bf R}^0\\
{\bf t} &= \exp( {\bm \xi}_R^{\wedge} ) {\bf t}^0 + {\bm \xi}_t \approx ( {\bf I} + {\bm \xi}_R^{\wedge} ) {\bf t}^0 + {\bm \xi}_t.
\label{eq:app}
\end{aligned}
\end{equation}
Then, the residual error is formulated as:
\begin{equation}
\begin{aligned}
{\bf r}_j &= {\cal T} {\bf v}_j - {\bm \mu}_j \\
&\approx  ( {\bf R}^0 + {\bm \xi}_R^{\wedge} {\bf R}^0 )  {\bf v}_j +  {\bf t}^0 + {\bm \xi}_R^{\wedge} {\bf t}^0 + {\bm \xi}_t - {\bm \mu}_j \\
&= {\bm \xi}_R^{\wedge} ({\bf R}^0 {\bf v}_j ) + {\bm \xi}_R^{\wedge} {\bf t}^0 +  {\bm \xi}_t + {\bf r}_j^0
\end{aligned}
\end{equation}
Given the property depicted in Eq. (\ref{eq:pro}), the residual error is converted into:
\begin{equation}
\begin{aligned}
{{\bf{r}}_j}  = 
\begin{bmatrix}
-( {\bf R}^0 {\bf v}_j + {\bf t}^0)^\wedge & {\bf I} 
\end{bmatrix}
{\bm \xi}  + {\bf r}_j^0 =  {\bf H}_i {\bm \xi} + {\bf r}_j^0. 
\end{aligned}
\end{equation}
By replacing ${{\bf{r}}_j}$ with term ${\bf H}_i {\bm \xi} + {\bf r}_j^0$, Eq. (\ref{eq:objf}) can be converted into a quadratic program (QP) problem depicted in Eq. (\ref{eq:qp}), which is easy to be solved.}

Obviously, the objective function of QP problem is a local approximation to the original objective function, thus ${\bf H}$ is Hessian matrix and ${\bf{b}}$ is the gradient vector. The convergence of above optimization behaves like a second-order optimization, which is faster than typical gradient-based methods. And the optimal perturbation is calculated as follows: 
\begin{equation}
{\bm \xi}^* =  - {\bf H}^{\dagger}  {\bf b}, 
\label{eq:solu}
\end{equation}
where $(\cdot)^\dagger$ denotes the pseudo-inverse operator. 
Finally, the rigid transformation can be updated by Eq. (\ref{eq:pertu}), which is the optimal solution of Eq. (\ref{eq:objf}). For each point set, the Lie algebra solver can be sequentially utilized to optimize its rigid transformation.

\subsection{Implementation} 
As a local convergent method, 3DMNDT requires to be provided with initial rigid transformations. Given data points with arbitrary orientation, we utilize the coarse registration method  \cite{zhu2016automatic} to estimate initial rigid transformation for multi-view registration. Different from some previous methods, this method should be additionally initialized with $K$ cluster centroids. Therefore, we uniformly sample $K$ data points from all initial aligned point sets and take them as the initial centroids to start iteration.
To obtain desired results, the proposed method should alternately operate cluster update, ND computation,  and transformation optimization until all rigid transformations have no significant change or the iteration number $h$ reaches the maximum value $H$. 

Based on the above description, the proposed method is summarized in Algorithm 1, where we set $H=300$. In the proposed method, there is a free parameter $K$, which requires to be predefined. Frankly speaking, it is difficult to set its optimal value. However, we can set the averaging number of data points contained in each cluster, which can indirectly determine the cluster number. Specifically, we set $(6+M)$ to be the averaging number of data points due to two reasons. First, this setting can ensure that most clusters contain at least 6 data points and reduce the invalid clusters, even when the number of point sets is small. Second, the cluster number remains stable and will not increase with the increase of the point set number.

	\begin{algorithm}[!t]
	    \caption{3DMNDT method}
	    \label{Algorithm_1}
	    \renewcommand{\algorithmicrequire}{\textbf{Input:}}
        \renewcommand{\algorithmicensure}{\textbf{Output:}}
	   \begin{algorithmic}[1]
	        \REQUIRE Point sets ${\mathbb V} = \{ {{\bf V}_i}\} _{i = 1}^M$, initial transformations ${\mathbb T}^0$.
	        \STATE Uniformly initialize ${\bm \mu}^0_k$, set $h=0, K = M + 6$;
	        \REPEAT   
            \STATE $h=h+1$;
            \STATE Update cluster $c^h(i,j)$ by Eq. (\ref{eq:corr1});
            % \FOR {$\left( k=1:K \right)$}  
            \STATE Compute NDT $( {\bm \mu}_k^h, {\bm \Omega}_k^h)$ by Eqs. (\ref{eq:mu}) and (\ref{eq:inf}). 
            % \ENDFOR
            % \STATE  Discard invalid data points;
            \FOR {$\left( i=2:M \right)$}  
            \STATE Optimize the perturbation ${{\bm{\xi }}_i ^{*, h}}$ by Eq. (\ref{eq:solu});
            \STATE Update the transformation ${\cal T}_i^h$ by Eq (\ref{eq:pertu}); 
            \ENDFOR
            \STATE Compute likelihood $L({\bm \Theta}^h)$ by Eq. (\ref{eq:LLH}). 
            \UNTIL{($\left| {L({ {\bm \Theta}^h}){\rm{ - }}L({  {\bm \Theta}^{h - 1}})} \right| < \varepsilon$) or $\left( h > H\right)$}
	  \end{algorithmic}
	\end{algorithm}    

\begin{table}[!t]
\centering
\renewcommand\arraystretch{1.5}
\caption{The total computation complexity of the proposed method in each operation}
\begin{tabular}{cc}% |c|c|
\toprule
\textbf{Operation} & \textbf{Complexity} \\
\midrule
Build $k$-d tree & $O(HK\lg K)$\\
Build correspondence & $O(HN\lg K)$ \\
Compute NDT & $O(HN)$\\
Optimize transformation & $O(HN)$\\
\bottomrule 
\end{tabular}
\label{Tab:Com}
\end{table}

\subsection{Computation complexity}   
This section analyzes the computation complexity of 3DMNDT. Before analysis, we restate that there are $M$ point sets being registered, where the $i$-th point set contains $N_{i}$ data points and the total data points is $N=\Sigma_{i=1}^{M}N_{i}$. Besides, these data points desire to be divided into $K$ clusters and the number of valid clusters is $K^{'}<K$. For the convenience of analysis, we assume the iteration number of this method is $H$. In each iteration, three operations are included to optimize one rigid transformation.

\textbf{Update cluster}. Before updating cluster, it requires to build $k$-d tree for $K$ cluster centroids, which leads to the total complexity of $O(HK\lg K)$ for $H$ iterations. Then, each data point should be assigned to one cluster by the NN search, which leads to the total complexity of $O(HN\lg K)$ for $H$ iterations.

\textbf{Compute NDT}. After updating cluster, the NDT should be computed for each cluster. Since each data point is only assigned to one cluster and each cluster utilizes all its data points to compute one NDT, the complexity is $O(N)$ in each iteration. For $H$ iterations, the total complexity is $O(HN)$ 

\textbf{Optimize rigid transformation}. The proposed method utilizes $N_i^{'} \approx N_i $ point-centroid pairs to optimize the $i$th rigid transformation. To estimate $M$ rigid transformations, the total complexity is $O(HN)$ for $H$ iterations.

Accordingly, Table \ref{Tab:Com} displays the total computation complexity of the proposed method. As shown in Table \ref{Tab:Com}, the highest complexity operation is linear proportion to $N$, $H$ and $lg K$. Usually, the cluster number $K$ is much smaller than the point number $N$. What's more, $\lg K$ is much smaller than the cluster number $K$ . Therefore, the proposed method is efficient.

\section{Experiment results}   
In this section, we evaluate the proposed 3DMNDT method on various benchmark data sets, which include: (1) \textbf{Object data sets.} Four data sets are downloaded from the Stanford 3D Scanning Repository and two data sets are provided by Torsello. Each of them was acquired from one object model in multiple view-points. (2) \textbf{Environment data set.} Gazebo data set (Courtesy of Francois Pomerleau) is acquired from outdoor environment and it is provided for robot simultaneous localization and mapping (SLAM). Table II displays the scan number and unit for each data set.
In addition to scans, the ground truth of rigid transformations are also provided to evaluate the performance of registration method. For efficiency, each data set is uniformly down-sampled to around 2000 points per scan.

To illustrate its performance, we compare 3DMNDT with a number of baselines:

(1) \textbf{MATrICP \cite{govindu2013averaging}.} This method utilizes the motion averaging algorithm to recover global motions from a set of relative motions, which are estimated by the TrICP algorithm.

(2) \textbf{JRMPC \cite{evangelidis2017joint}.} This method supposes that all data points are drawn from a central GMM and utilizes the EM algorithm to estimate the GMM components as well as rigid transformations.

(3) \textbf{TMM \cite{evangelidis2017joint}.} This method supposes that all data points are drawn from a central Student’s t-mixture model (TMM) and utilizes the EM algorithm to estimate the TMM components as well as rigid transformations.

(4) \textbf{LRS \cite{arrigoni2018robust}} This method concatenates available relative motions into one matrix, which is decomposed by the low rank and sparse (LRS) decomposition algorithm to recover global motions for multi-view registration.

(5) \textbf{K-means \cite{zhu2019efficient}.} This method casts multi-view registration into clustering problem and extends the k-means clustering algorithm to achieve multi-view registration.

(6) \textbf{EMPMR \cite{zhu2020registration}.} This method assumes that each data point is generated by one unique GMM, where all Gaussian centroids are efficiently searched from each other scans by the $k$-d based method. 

 \begin{table}[!t]
\setlength{\tabcolsep}{0.9mm}
\centering
\caption{Details of bench mark data sets utilized in the experiment.}
\begin{tabular}{ccccccccc}
 \toprule
{Dataset} & Angel & Armadillo& Bunny & Buddha & Dragon & Hand &  Gazebo \\
\midrule
Scans & 36 & 12 & 10 & 15 & 15 & 36 & 32 \\
\midrule
Unit &dm &mm &mm &mm &mm&dm &m \\ 
\bottomrule
\end{tabular}
\label{Tab:Data}
\end{table}  

Experimental results are reported in the form of rotation error and translation error defined as:
\begin{equation}
e_{\bf R} = \frac{1}{M} \sum_{i = 1}^M {\cos^{-1} ( {\rm Tr} ( {\bf R}_{m, i}  {\bf R}_{g, i}^\top ) - 1 )\over 2},
\end{equation}
\begin{equation}
e_{t}=\frac{1}{M}\sum_{i=1}^M\| {\bf t}_{m,i} - {\bf t}_{g,i}\|_{2},
\end{equation}
where ${\rm Tr}(\cdot)$ denotes the trace operator on a square matrix, ${\rm cos}^{-1}: [-1, 1] \rightarrow [0, \pi]$ denotes the inverse cosine function, $\{ {\bf R}_{g,i},{\bf t}_{g,i}\}$ and $\{{\bf R}_{m,i}, {\bf t}_{m,i}\}$ denote the ground truth and the estimated one of the $i$-th rigid transformation, respectively. Therefore, the unit of $e_{\bf R}$ is radian and the unit of $e_{t}$ is similar to the unit of each data set. All compared methods are implemented on Matlab without any extra library and utilize the $k$-d tree method to search the NN. Experiments are carried out on a quad-core 3.6 GHz computer with 8 GB of memory. 

\begin{table}[!t]
\centering
\setlength{\tabcolsep}{0.7mm}
\caption{Error comparison of different NDT variants, where bold numbers denote the best performance of each data set.}
\begin{tabular}{cccccccccc} 
\toprule
Method                   &       & Angel            & Armadillo          & Bunny           & Buddha          & Dragon            & Hand       &    Gazebo   \\
\midrule
\multirow{3}{*}{Initial}  & $e_{\bf R} $ &0.0221 &0.0234 &0.0239 &0.0262 &0.0251  &0.0282  &0.0241\\
                                  & $e_t$ &2.0388 &2.5333 &2.1260 &1.6535 &1.5216  &0.4945 & 0.1069\\
\midrule
\multirow{3}{*}{NDTO}   & $e_{\bf R} $ & 0.0258&0.0057	&0.0113&0.0232	&0.0349	&0.0214	&0.0422 \\
                        & $e_t$  &1.4885 &2.3317	&1.0811&4.7053	&6.5298&0.6159	&0.5948\\ 
\midrule
\multirow{3}{*}{NDTL}   & $e_{\bf R} $ &0.0422  &0.0054 &0.0041	&0.0095 &0.0084	&0.0174	&0.0650	\\
                        & $e_t$ &2.1972  &0.8756 &0.6865	&0.9500&1.6755	&1.7976	&0.3092	\\ 
\midrule
    \multirow{3}{*}{3DMNDT}  & $e_{\bf R} $ & \textbf{0.0008}	&\textbf{0.0049}	&\textbf{0.0024}	&\textbf{0.0073}	&\textbf{0.0103}	&\textbf{0.0014}	&\textbf{0.0082}\\
                                  & $e_t$ &\textbf{0.1287}	&\textbf{0.6688}	&\textbf{0.2426}	&\textbf{1.0712}	&\textbf{1.3531}	&\textbf{0.0876}	&\textbf{0.0188}\\ 
\bottomrule
\end{tabular}
\label{Tab:Val}
\end{table}

\subsection{Validation of clustering} 
To validate the proposed 3DMNDT, we compared it with two variants of NDT algorithm, where NDTO sequentially registers each scan to other aligned scans by original NDT and NDTL replaces the Newton-based solver with the Lie algebra based solver in NDTO to achieve registration. Experimental results are reported in the form of registration errors, which are displayed in Table \ref{Tab:Val}. As shown in Table \ref{Tab:Val}, 3DMNDT can obtain more accurate registration results among three competed methods. 

\begin{figure}[!t]
\centering
%\subfigure[]{\includegraphics[scale=0.9]{Fig5a.pdf}}
%\vfill
\subfigure[]{\includegraphics[scale=0.39]{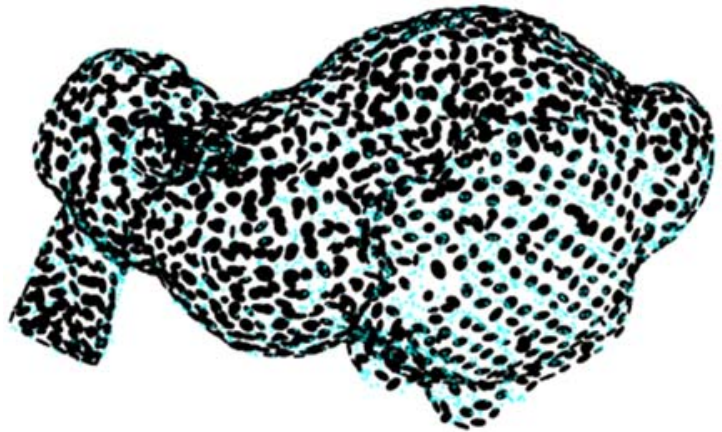}}
\subfigure[]{\includegraphics[scale=0.39]{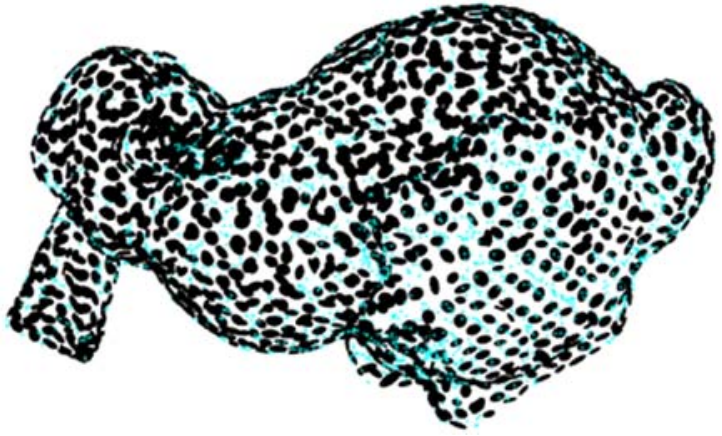}}
\subfigure[]{\includegraphics[scale=0.39]{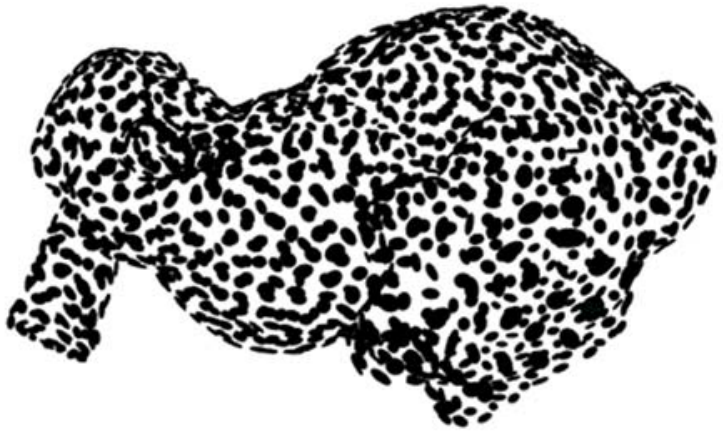}}
\caption{Illustration of NDTs for different methods, where cyan points denote these data points belong to invalid clusters. (a) Results of NDTO. (b) Result of NDTL. (c) Result of 3DMNDT.}
\label{Fig:NDT}
\end{figure}

\begin{figure}[!t]
\centering
\includegraphics[scale=0.45]{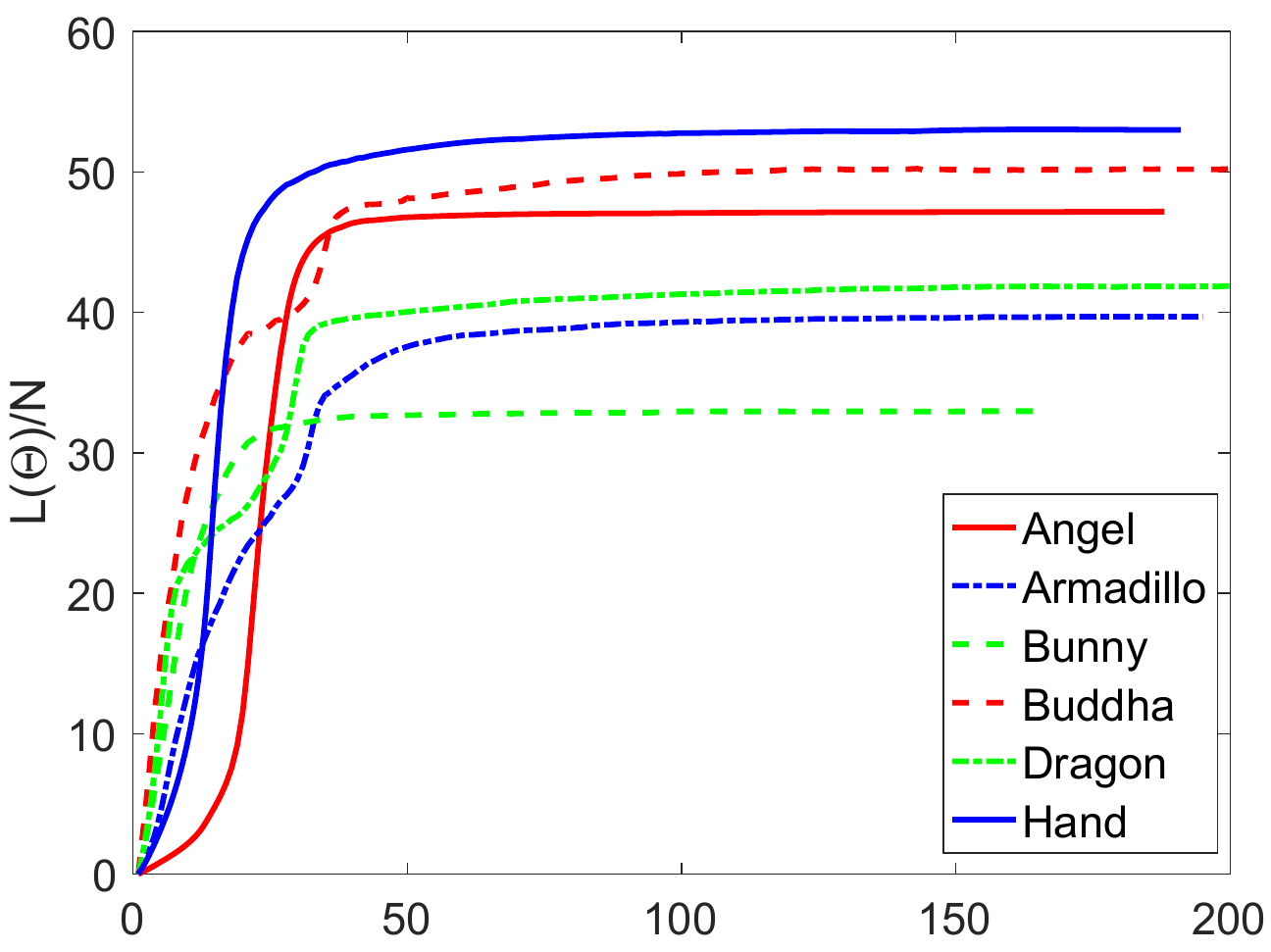}
\caption{ Convergence results of 3DNDT tested on object data sets. For view convenience, we display the log-likelihood value subtracted by the initial value for each data set.}
\label{Fig:Conv}
\end{figure}

Usually, data points are non-uniformly distributed in the occupied space. As shown in Fig. \ref{Fig:NDT}, different division strategies lead to different NDT results. In NDTO and NDTL, the occupied space are divided into regular cells, which inevitably lead to unbalanced division. Specifically, some cells may only contain a small number of data points, i.e. smaller than 6. To avoid singular covariance, data points located in these cells should be neglected to compute NDT and optimize rigid transformations, which inevitably reduces the performance of multi-view registration. Accordingly, both NDTO and NDTL are difficult to achieve promising results for multi-view registration. What's more, the Newton-based solver in NDTO is difficult to be implemented and debugged due to two reasons: Firstly, the Jacobian and Hessian matrices frequently generate many pages of dense algebra that should be converted into code, which is a tedious, difficult and error-prone process. Secondly, the Newton-based solver also involves complex line search, which requires hundreds lines of code. 

While, the k-means clustering is able to avoid unbalanced division and guarantee that most clusters contain a certain number of points. Since 3DMNDT utilizes k-means clustering, most of data points are utilized to compute NDTs and optimize rigid transformations. Compared with the Newton's solver, the Lie algebra solver is a more direct and explicit mechanism for the optimization of 3D rigid transformation. Fig. \ref{Fig:Conv} also displays the convergence of 3DMNDT with respect to iterations. As shown in Fig. \ref{Fig:Conv}, the log-likelihood value increases with the increase of iterations. Thanks for K-means clustering and the Lie algebra solver, which enable 3DMNDT to achieve promising registration results within acceptable iterations.

\subsection{Object data sets} 
To illustrate its registration performance, 3DMNDT is tested on object data sets and compared with a number of baselines on accuracy, efficiency, and robustness.

\begin{table}[!t]
\centering
\setlength{\tabcolsep}{1.2mm}
\renewcommand\arraystretch{1.0}
\caption{Error comparison of all competed methods, where bold numbers denote the best performance of each data set.}
\begin{tabular}{cccccccc}
\toprule 
Method                   &       & Angel              & Armadillo          & Bunny             & Buddha           & Dragon           & Hand            \\
\midrule
\multirow{2}{*}{Initial} & $e_{\bf R}$ &0.0221 &0.0234 &0.0239 &0.0262 &0.0251  &0.0282 \\
                                  & $e_t$ &2.0388 &2.5333 &2.1260 &1.6535 &1.5216  &0.4945  \\
\midrule
\multirow{2}{*}{MATrICP} & $e_{\bf R} $ &0.0103 &0.0237 &0.0088 &0.0085 &0.0116  &0.0262 \\
                                  & $e_t$ &2.8651 &4.2915 &0.7181 &\textbf{0.8581} &1.1037  &1.3390  \\
\midrule
\multirow{2}{*}{JRMPC}& $e_{\bf R} $ &0.0065 &0.0121 &0.0166 &0.0161 & 0.0154 & 0.0044 \\
                                & $e_t$ &1.9816 &3.0651 &1.7571 &0.9629 &1.7466  &0.7069 \\
\midrule
\multirow{2}{*}{TMM} & $e_{\bf R} $&0.0062 &0.0061 &0.0080 & 0.0106& 0.0154 &0.0072 \\
                                  & $e_t$ &2.5197 &1.9646 &0.7031 &1.2137 &\textbf{1.0689}  &0.7682  \\
\midrule
\multirow{2}{*}{LRS}  & $e_{\bf R} $ &0.0087 &0.0410 &0.0125 &0.0181 & 0.0162 &0.0265 \\
                                  & $e_t$ &2.0633 &9.4194 &0.9586 &1.0597 &1.5358  &1.3825 \\ 
\midrule
\multirow{2}{*}{K-means}  & $e_{\bf R} $ &0.0056 &0.0066 &0.0100 &0.0147 &0.0134  & 0.0059\\
                                  & $e_t$ &0.9789 &2.3207 &1.6088 & 1.0276 &1.6140 &0.4615 \\ 
                                  \hline 
    \multirow{2}{*}{EMPMR}  & $e_{\bf R} $   &0.0012 &0.0052 &0.0035 &\textbf{0.0067} &\textbf{0.0090} &\textbf{0.0014} \\
                                  & $e_t$ &0.2161 &0.7145 &0.3439 &1.1023 &1.1433  &0.0887 \\ 
\midrule
    \multirow{2}{*}{3DMNDT}  & $e_{\bf R} $   &\textbf{0.0008} &\textbf{0.0049} &\textbf{0.0024} &0.0073 &0.0103 &0.0015 \\
                                  & $e_t$ &\textbf{0.1287} &\textbf{0.6688} &\textbf{0.2426} &1.0712 &1.3531  &\textbf{0.0876} \\
\bottomrule
\end{tabular}
\label{Tab:Acc}
\end{table}

\begin{figure*}[!t]
\centering
\includegraphics[scale= 0.95]{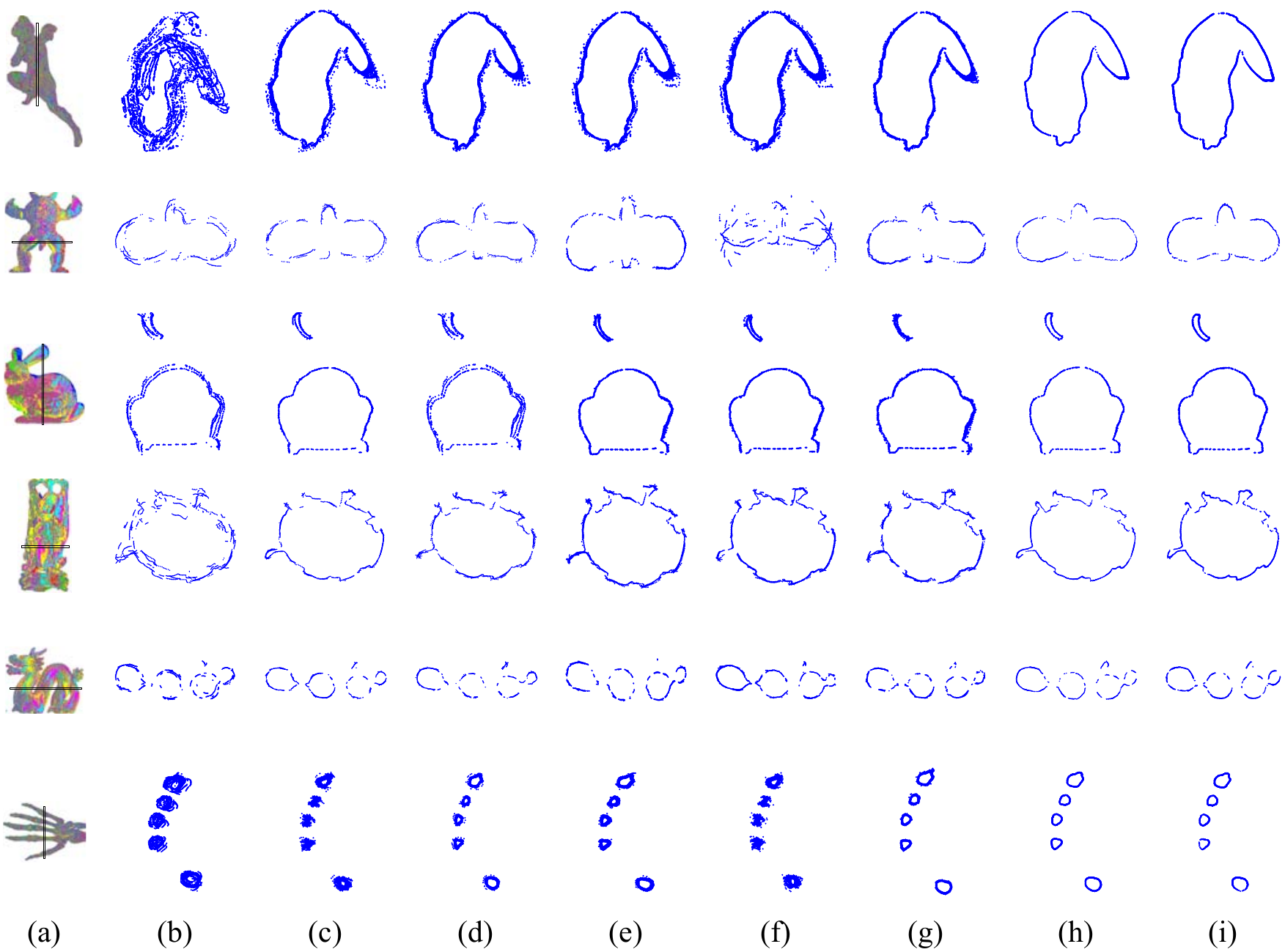}
\caption{Multi-view registration results in the form of cross-section, where corresponding regions are illustrated on the aligned 3D models and thin curves denote good registration results. (a) Aligned 3D models. (b) Initial results. (c) Results MATrICP. (d) Results of JRMPC. (e) Results of TMM. (f) Results of LRS. (g) Results of K-means. (h) Results of EMPMR. (i) Results of 3DMNDT.}
\label{Fig:Cros}
\end{figure*}

\subsubsection{Accuracy}   
To compare registration accuracy, these registration methods are tested on six object data sets, experimental results are reported in registration errors. Table \ref{Tab:Acc} illustrates registration errors of all compared methods tested on six object data sets. For more intuitive comparison, Fig. \ref{Fig:Cros} displays all registration results in the form of cross-section. As shown in Table \ref{Tab:Acc} and Fig. \ref{Fig:Cros}, 3DMNDT and EMPMR are able to obtain the most promising registration results for most of data sets.

For the multi-view registration, both MATrICP and LRS recover global motions from a set of available relative motions, which are estimated by the TrICP algorithm. Given accurate relative motions, these two methods may obtain promising registration results.
Since the TrICP algorithm is not error free, relative motions inevitably contain errors, which reduce the accuracy of multi-view registration. What's more, MATrICP is sensitive to unreliable relative motions, even one unreliable relative motions will lead to its registration failure, e.g. the Armadillo data set. Meanwhile, LRS is dependent on the ratio of available relative motions and low ratio will lead to registration failure, e.g. the Armadillo data set. As probabilistic methods, both JRMPC and TMM are expected to obtain very accurate registration results. However, this is not the case. Because these two methods require to estimate 
massive mixture model parameters as well as rigid transformations, which makes them to be easily trapped into local minimum. While, EMPMR only requires to estimate rigid transformations as well as one Gaussian variance, so it is more likely to obtain promising registration results. 

Although both K-means and 3DMNDT utilize k-means clustering to achieve multi-view registration, their performance is totally different. In K-means, all data points in each cluster are approximated by one centroid, which inevitably leads to information loss. As K-means directly aligns each point set to all cluster centroids for registration, it is difficult to obtain promising results. While, 3DMNDT computes NDT for each cluster and achieves multi-view registration by maximizing NDT-based likelihood function. Since NDT contains statistical information of each cluster, 3DMNDT is more likely to obtain accurate registration results.

\subsubsection{Efficiency} 
\begin{figure}[!t]
\centering
\includegraphics[scale=0.55]{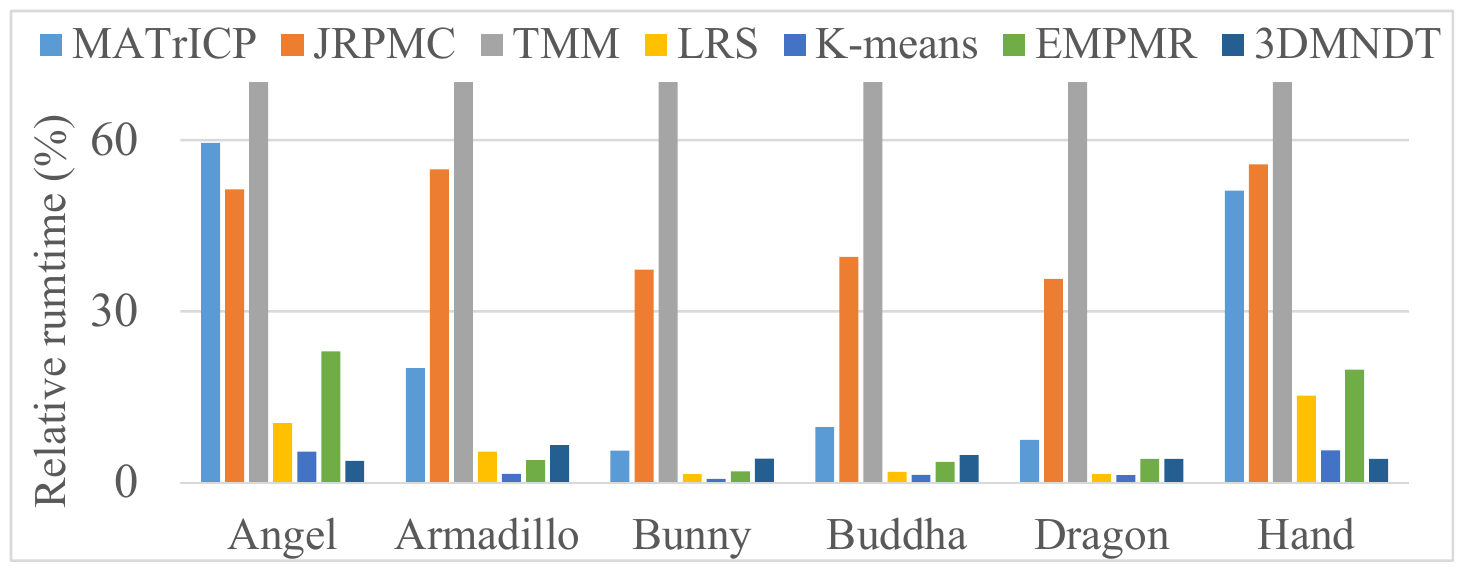}
\caption{Efficiency comparison over 10 independent tests. For view convenience, we display relative run time for all competed methods, where TMM is most time-consuming method with (17.2717, 5.4772, 6.5912, 9.276, 10.4413, 15.9024) minutes corresponding to 100\% run time of each data set.}
\label{Fig:Time}
\end{figure}

To compare registration efficiency, these registration methods are tested on six object data sets. Experimental results are reported in the form of averaging run time over 30 independent tests. Fig. \ref{Fig:Time} illustrates the efficiency comparison of these registration methods tested on object data sets. As shown in Fig. \ref{Fig:Time}, K-means is the most efficient one among all registration methods and 3DMNDT is comparable with K-means especially for large data sets. Besides, TMM and JRPMC are more time-consuming than other registration methods.

 \begin{figure*}[!t]
\centering
\subfigure[]{\includegraphics[scale=0.345]{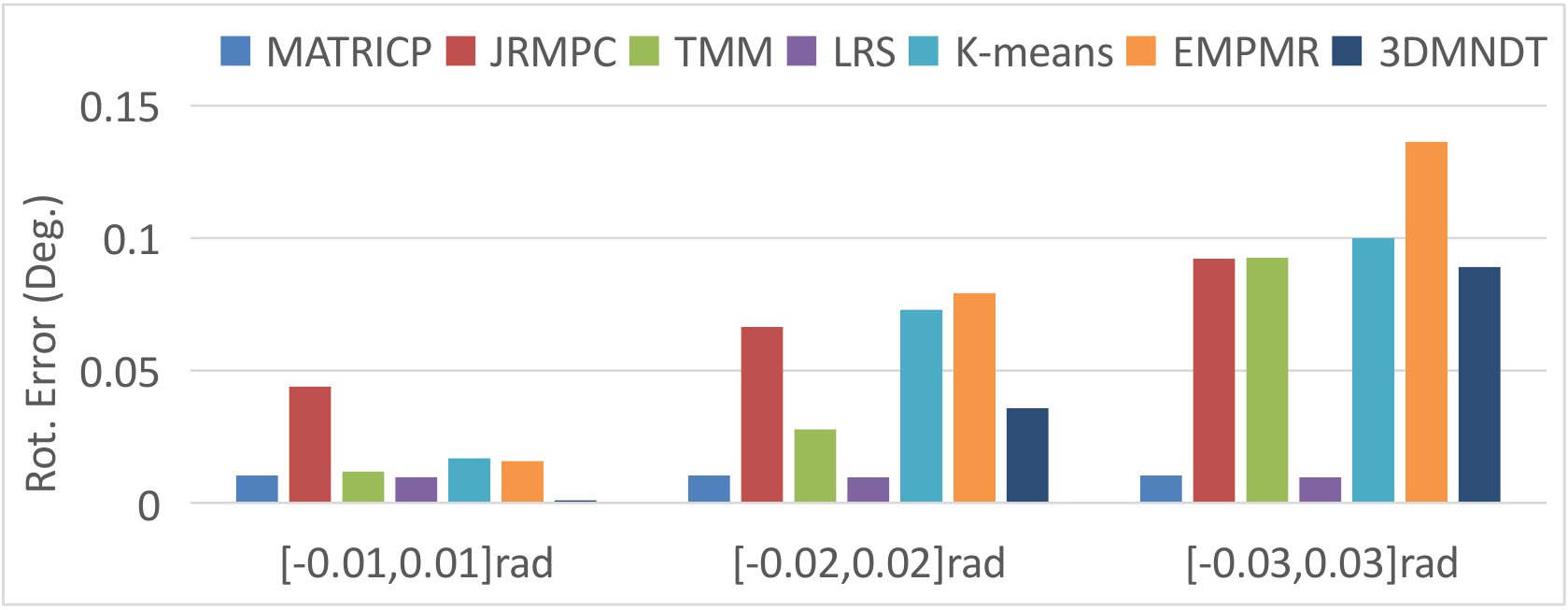}}
\subfigure[]{\includegraphics[scale=0.345]{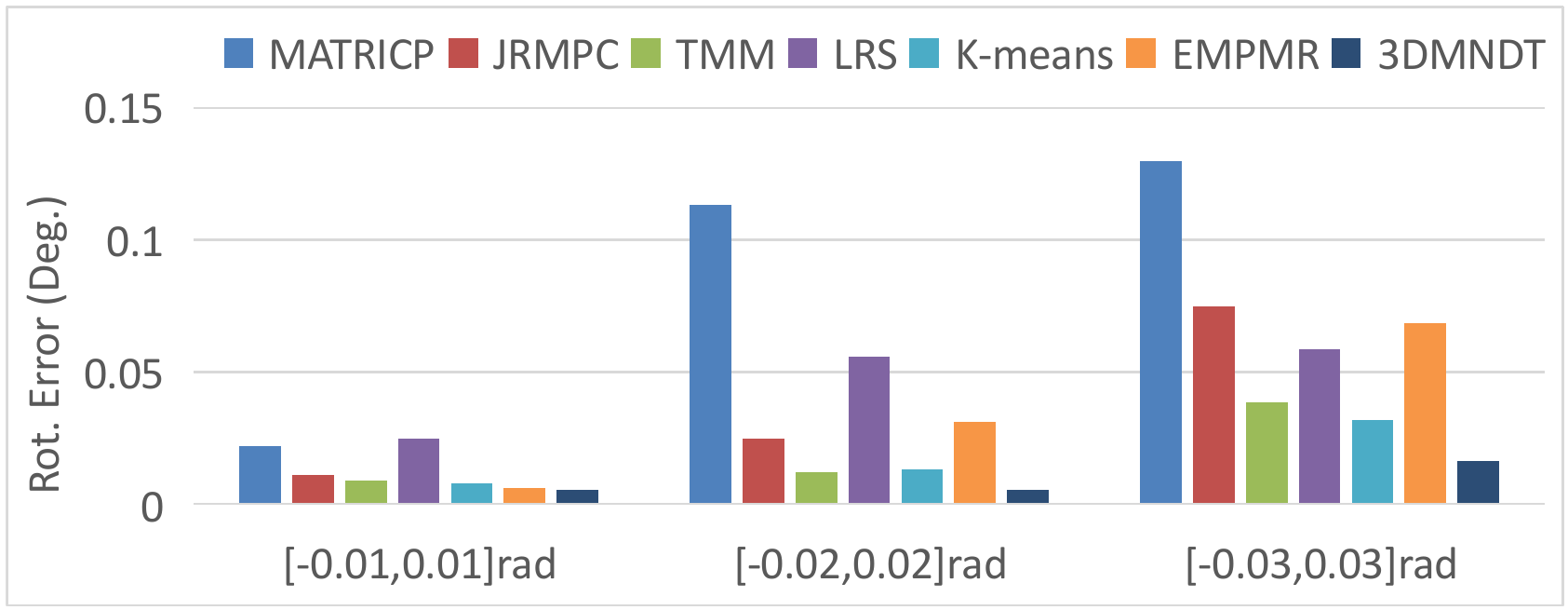}}
\subfigure[]{\includegraphics[scale=0.345]{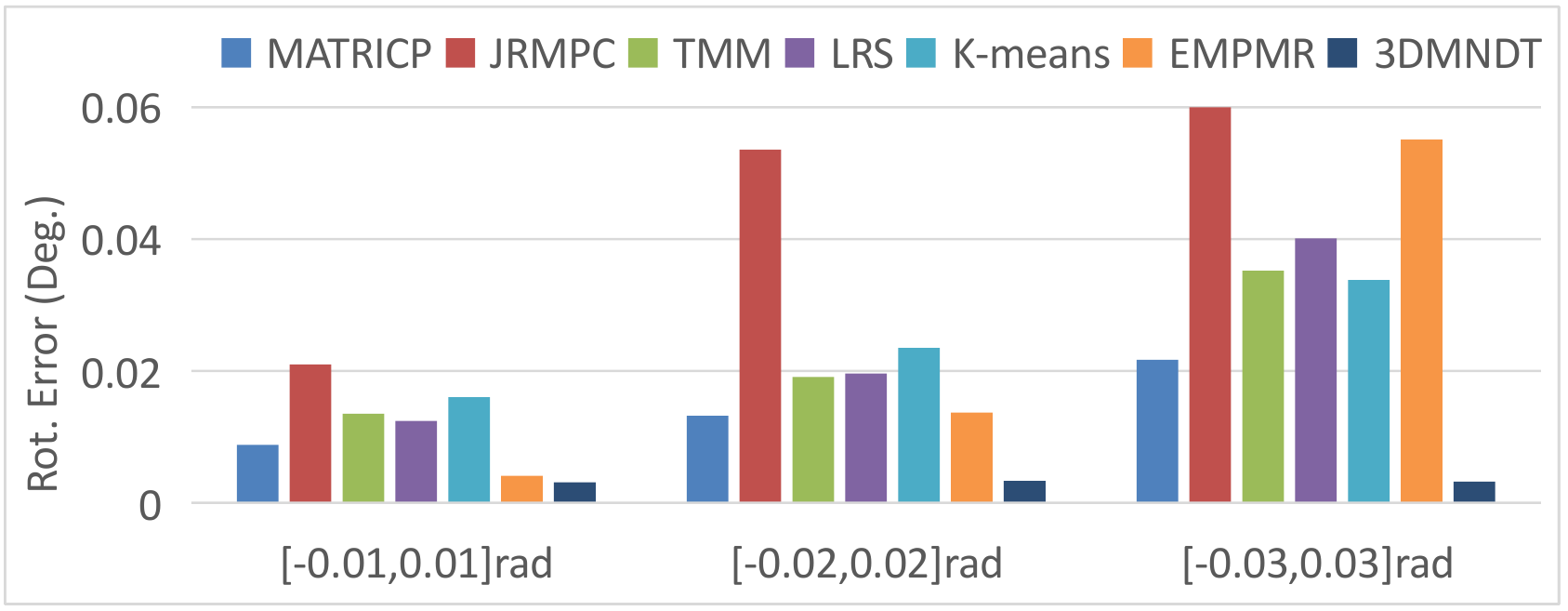}}
\subfigure[]{\includegraphics[scale=0.345]{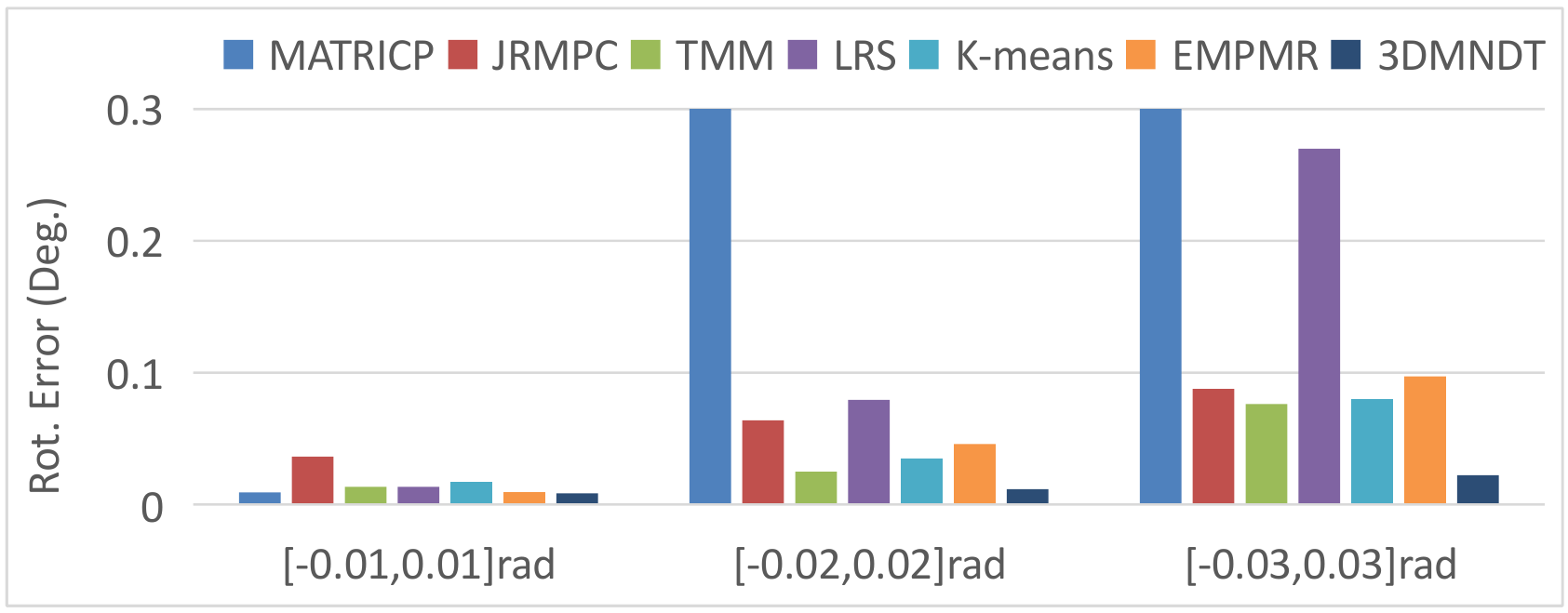}}
\subfigure[]{\includegraphics[scale=0.345]{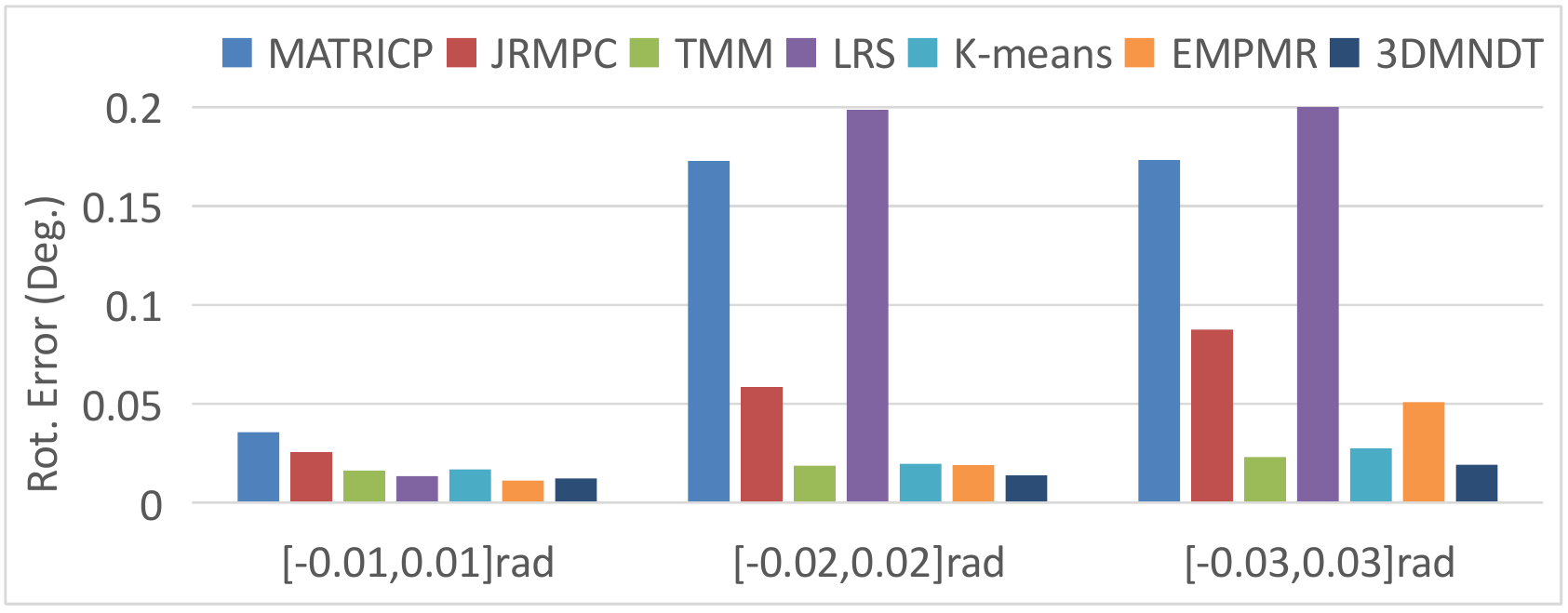}}
\subfigure[]{\includegraphics[scale=0.345]{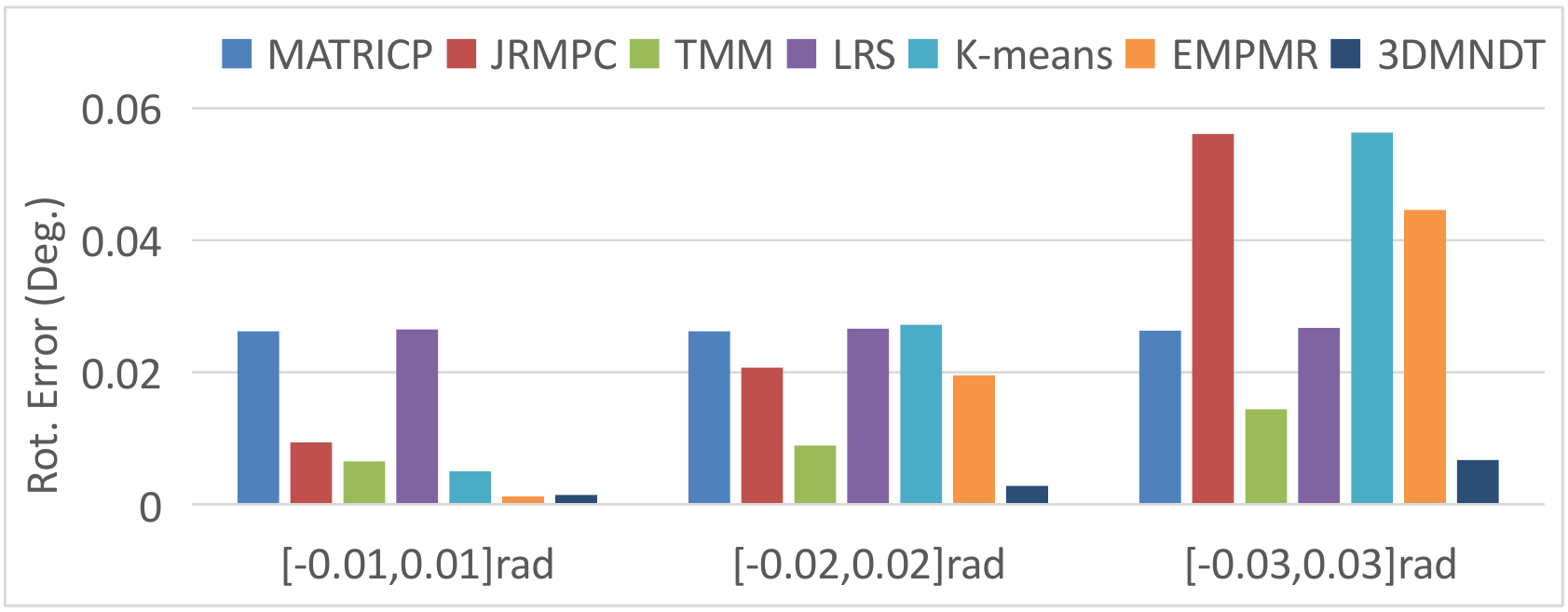}}
\caption{Comparison of averaging rotation errors for each object data set. (a) Angle. (b) Armadillo. (c) Bunny. (d) Buddha. (e) Dragon. (f) Hand.}
\label{Fig:Rot}
\end{figure*}

 \begin{figure*}[!t]
\centering
\subfigure[]{\includegraphics[scale=0.345]{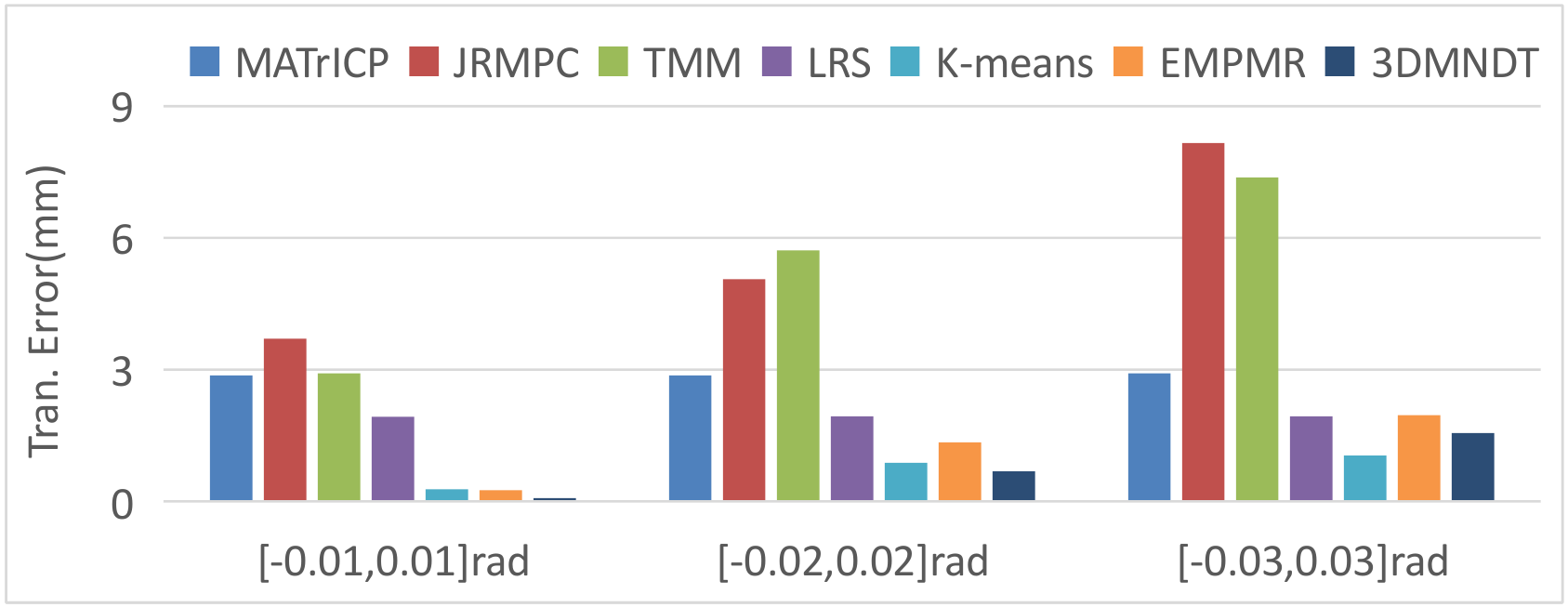}}
\subfigure[]{\includegraphics[scale=0.345]{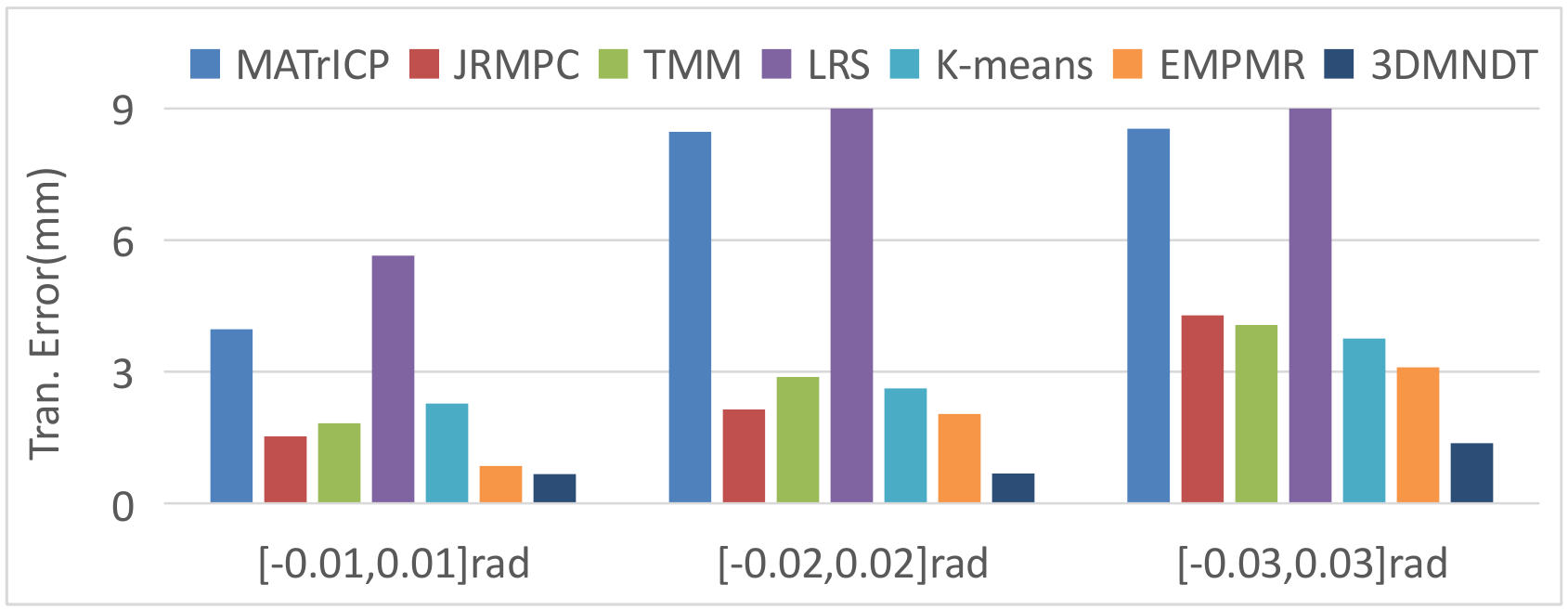}}
\subfigure[]{\includegraphics[scale=0.345]{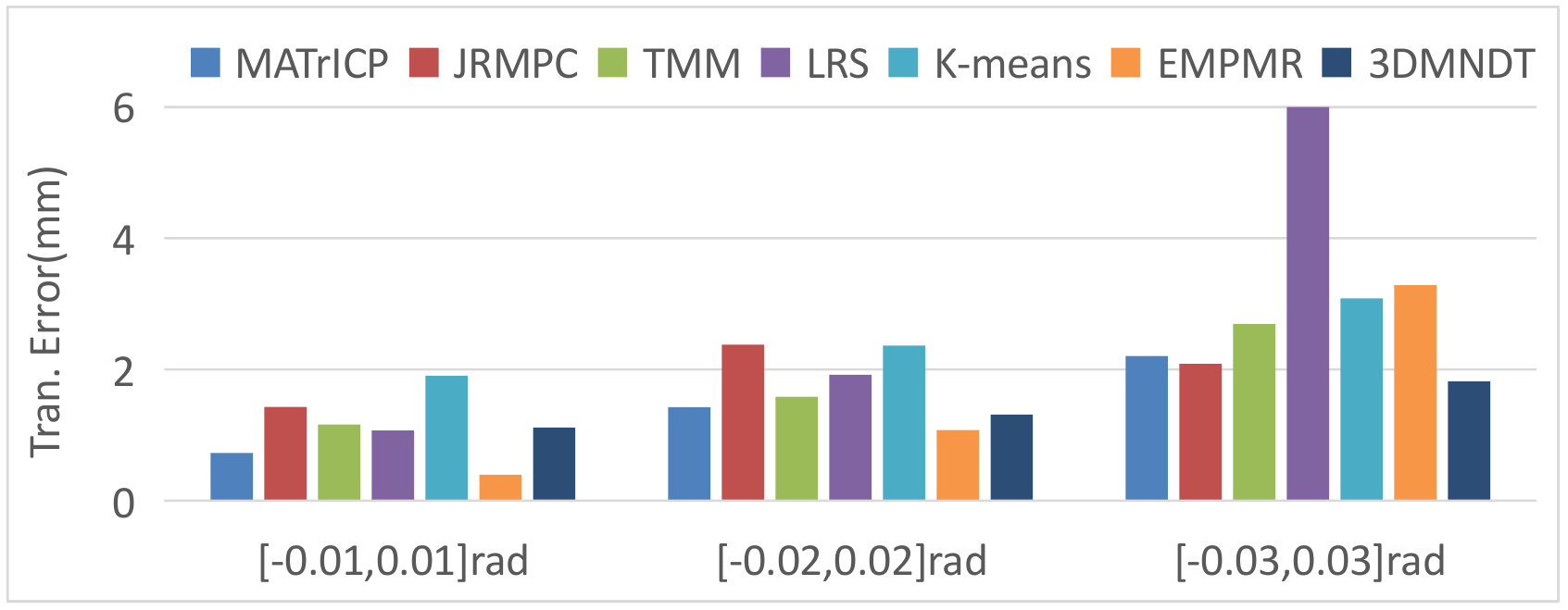}}
\subfigure[]{\includegraphics[scale=0.345]{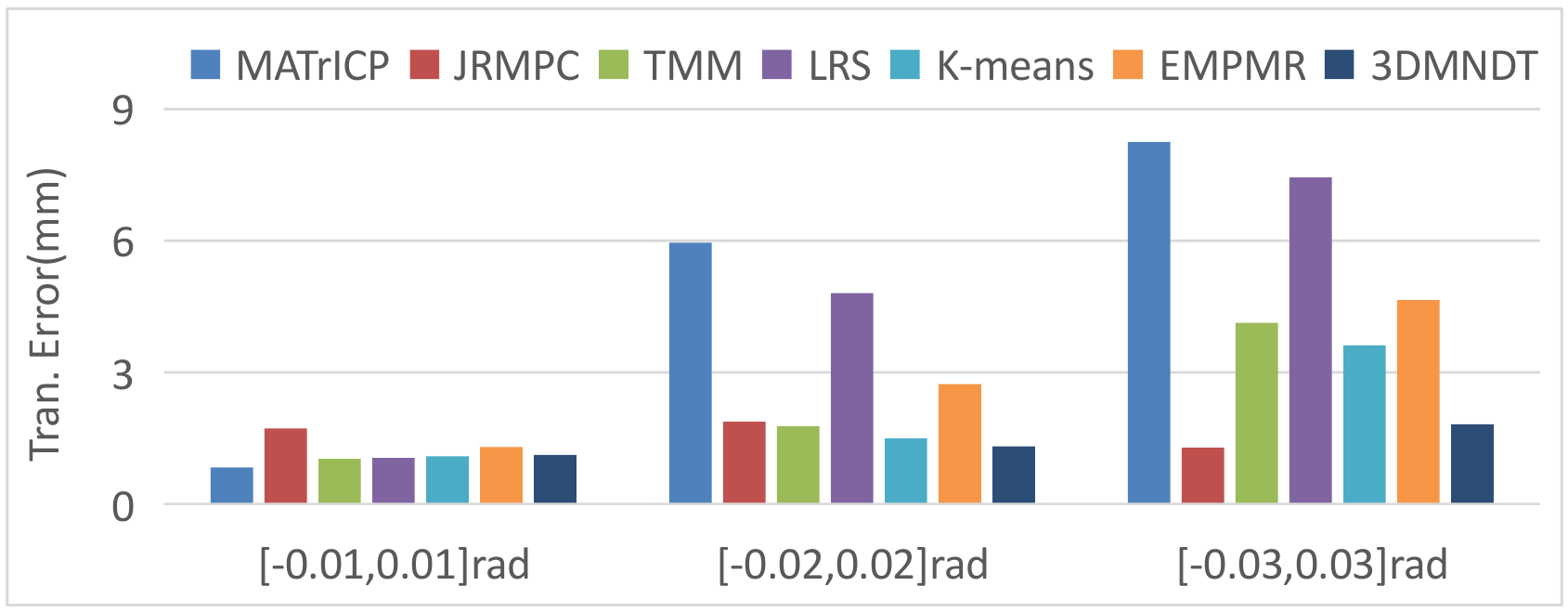}}
\subfigure[]{\includegraphics[scale=0.345]{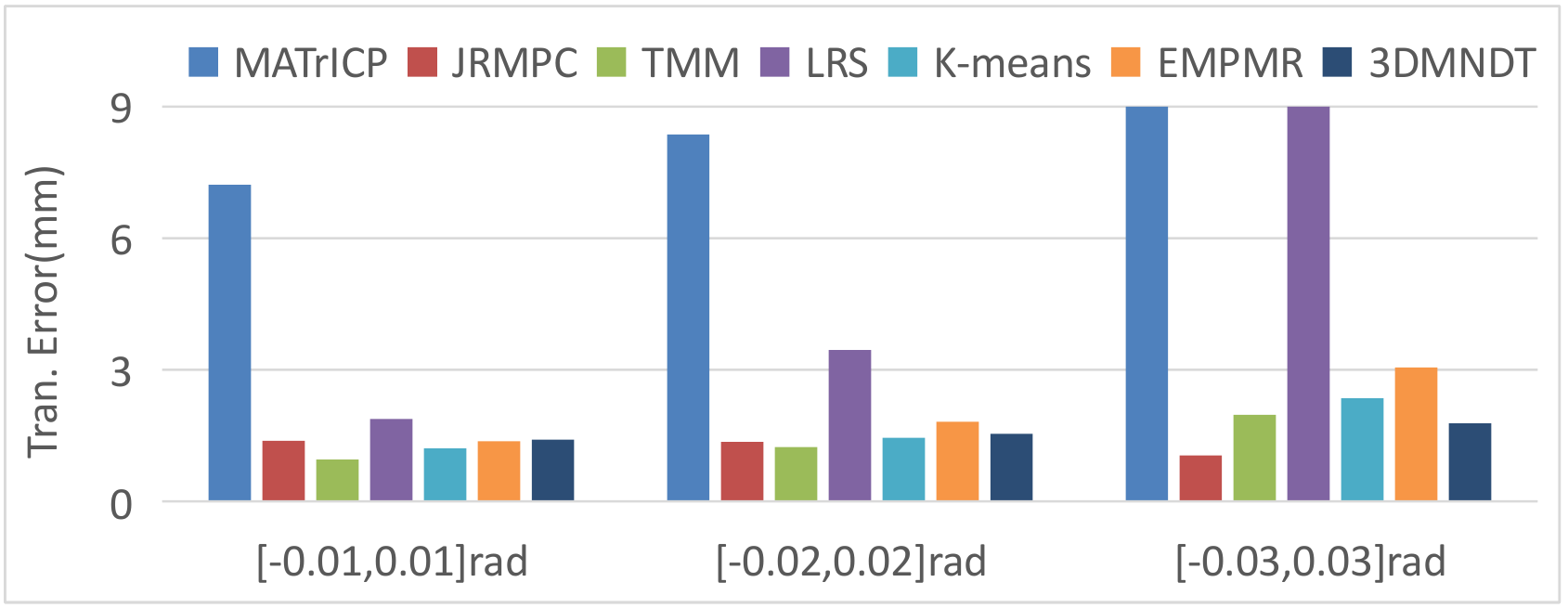}}
\subfigure[]{\includegraphics[scale=0.345]{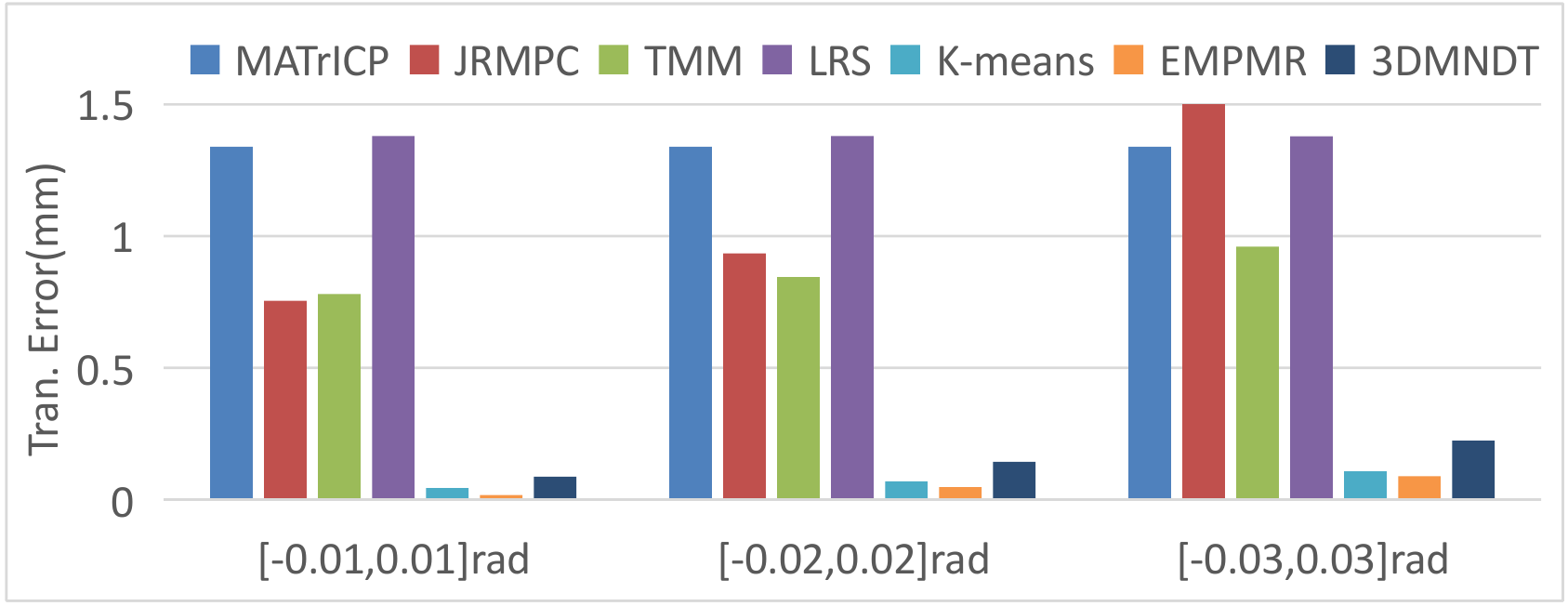}}
\caption{Comparison of averaging translation errors for each object data set. (a) Angle. (b) Armadillo. (c) Bunny. (d) Buddha. (e) Dragon. (f) Hand.}
\label{Fig:Trans}
\end{figure*}

Both TMM and JRPMC assume that all data points are generated from a central mixture model. To achieve multi-view registration, EM algorithm is utilized to estimate all mixture components as well as rigid transformations. Since each mixture component and rigid transformation are related to all data points, the estimation is very time-consuming. Compared with JRPMC, TMM requires to estimate another set of hidden variables. Therefore, TMM is the most time-consuming one among all compared methods. As aforementioned, both MATrICP and LRS recover global motions from available relative motions. Therefore, it requires to build point correspondences between most scan pairs. Since the convergence speed of MATrICP is slower than that of LRS, it is less efficient than LRS. With the increase of scan number, MATrICP even becomes less efficient than JRPMC. Besides, MATrICP is less efficient than EMPMR. Although EMPMR requires to build point correspondences between each scan pair, its convergence speed is faster than that of MATrICP. Accordingly, it is more efficient than MATrICP but a little less efficient than LRS.

Similar to K-means, 3DMNDT also utilizes K-means clustering algorithm to achieve multi-view registration. Specifically, 
K-means compute a centroid and minimize the sum function of point-centroid errors to optimize rigid transformations. While, 
3DMNDT requires to compute NDT and minimize the NDT-based likelihood function. Therefore, 3DMNDT is a little less efficient than K-means, but is more efficient than other registration methods, especially for large data sets.

\subsubsection{Robust to initialization} 

To illustrate its robustness to initialization, 3DMNDT as well as other registration methods are also tested on object data sets with different initial rigid transformations, where rotation angles are added with three levels of uniformly distributed noises. Considering randomness, each group of experiment is carried out by 20 independent tests. Experimental results are reported in the form of averaging registration errors. Figs. \ref{Fig:Rot} and \ref{Fig:Trans} demonstrate averaging rotation errors and translation error of all competed methods, respectively. Considering both rotation and translation errors, 3DMNDT is generally the most robust one to initialization among all competed methods. 

Both MATrICP and LRS recover global motions from a set of relative motions, which are estimated by the pair-wise registration algorithm. Given a set of reliable relative motions, MATrICP can obtain promising registration results. However, this method is sensitive to unreliable relative motions. Without good initialization, the probability of obtaining unreliable relative motions will increase, which  will lead to the failure of MATrICP. Although LRS is robust to unreliable relative motions, it is sensitive to the ratio of reliable relative motions. Under poor initialization, the number of reliable relative motions will be reduced, which will lead to the failure of LRS. Therefore, LRS is also sensitive to initialization. Since JRMPC and TMM requires to estimate massive model parameters, they are more likely trapped into local minimum, especially when the initialization is poor. Therefore, both of them are also sensitive to initialization.

Benefiting from data clustering, both K-means and 3DMNDT have wide convergent domains. However, K-means is difficult to obtain promising registration results due to information loss. Since 3DMNDT utilizes one NDT to represent data points of each cluster, it is more likely to obtain promising results due to less information loss. Thereby, 3DMNDT is the most robust one to initialization among all competed methods.

\begin{table*}[!t]
\centering
\setlength{\tabcolsep}{2mm}
\renewcommand\arraystretch{1.2}
\caption{Comparison of registration error (mean) for competed methods tested on six data sets perturbed by two level of Gaussian noises, where bold numbers denote the best performance. }
\begin{tabular}{cccccccc|cccccc}  % |c|c|c|c|c|c|c|c|c|c|c|c|c|c|
\toprule
\multirow{2}{*}{Method} &  \multirow{2}{*}{Error}    
& \multicolumn{6}{c|}{50dB}   &\multicolumn{6}{c}{25dB} \\
% \cline{3-14} 
& & Angel &Armadillo &Bunny &Buddha &Dragon &Hand & Angel &Armadillo &Bunny &Buddha &Dragon &Hand \\ 
% \cline{3-14} 
\midrule
\multirow{2}{*}{MATrICP}    & $e_{\bf R}$ & 0.0102 & 0.0204 & 0.0089 & 0.0086 & 0.0114 & 0.0261  & 0.0103  & 0.0204  & 0.0119  & 0.0123  & 0.0162  & 0.0369  \\
                            & $e_t$ & 2.8710 & 4.0146  & 0.6996  & \textbf{0.8424 }& \textbf{1.0464 } & 1.3417  & 2.8726  & 3.8764  &0.7047  &\textbf{ 0.8433 } & \textbf{1.0534 } & 1.3419 \\
\midrule
\multirow{2}{*}{JRMPC}    & $e_{\bf R}$  & 0.0062  & 0.0119  & 0.0146  & 0.0162  & 0.0153  & 0.0044  & 0.0062  & 0.0118  & 0.0165  & 0.0154  & 0.0157  & 0.0048 \\

                            & $e_t$ & 1.9069  & 3.0249  & 1.7402  & 0.9690  & 1.7823  & 0.7075 & 1.9628  & 3.0543 & 1.7404  & 0.9694  & 1.7923  & 0.7178 \\
                           
\midrule
\multirow{2}{*}{TMM}        & $e_{\bf R} $ &0.0061 &0.0060 &0.0084 & 0.0102 &0.0153 &0.0072 &0.0062 &0.0062 &0.0086 &0.0104 &0.0155 &0.0073\\
                            & $e_t$ &2.4665 &1.9844 &0.7487 &1.1904 &1.0662 &0.7792 &2.4865 &1.9913 &0.7626 &1.1950 &1.0686 &0.7802\\
\midrule
\multirow{2}{*}{LRS}         & $e_{\bf R}$ &0.0086 &0.0397 &0.0127 &0.0180  &0.0150 &0.0265  &0.0123 &0.0398 &0.0127 &0.0181 &0.0153 &0.0265\\
                            & $e_t$ &2.0599 &9.1893 &0.9882 &1.0443 &1.4367 &1.3803 &2.0559 &9.1627 &0.9905 &1.0551 &1.5105 &1.3817\\ 
\midrule
\multirow{2}{*}{K-means}         & $e_{\bf R}$ & 0.0056 & 0.0064 & 0.0104 & 0.0140 & 0.0141 & 0.0055 & 0.0057  & 0.0067  & 0.0107  & 0.0142  & 0.0151  & 0.0056 \\
                                  & $e_t$ & 1.0117  & 2.3448  & 1.6936  & 1.0684  & 1.3913  & 0.4674 & 1.0750  & 2.4698  & 1.7043  & 1.1212  & 1.5158 & 0.4690 \\
\midrule
\multirow{2}{*}{EMPMR}          &$e_{\bf R}$ &0.0012 &0.0052&0.0035 & \textbf{0.0068} &\textbf{ 0.0090}&\textbf{0.0014} &0.0012 &0.0055 &0.0037 &\textbf{0.0069}  &\textbf{0.0090}  &\textbf{0.0020} \\
                                  & $e_t$ & 0.2191& 0.7170 &0.3439 & 1.0972  &1.1179 &0.0839 & 0.2253 &0.7202 &0.3444  & 1.1001 & 1.1443&\textbf{0.1426}\\
\midrule
\multirow{2}{*}{3DMNDT}      & $e_{\bf R}$  &\textbf{0.0007}&\textbf{0.0046}&\textbf{0.0025 }&0.0076 &0.0108&\textbf{0.0014} &\textbf{0.0008} &\textbf{0.0046} &\textbf{0.0027} &0.0079 &0.0109 &0.0026\\

                             & $e_t$ &\textbf{0.1258} &\textbf{0.6551}&\textbf{0.2449}& 1.0899  &1.3507 & \textbf{0.0838}& \textbf{0.1488} &\textbf{0.6643} &\textbf{0.2519}  & 1.0956 &1.3729 &0.1567\\ 
\bottomrule
\end{tabular}
\label{Tab:50dB}
\end{table*}

 \begin{table*}[!t]
\setlength{\tabcolsep}{1.0mm}
\renewcommand\arraystretch{0.8}
\centering
\caption{Comparison of localization accuracy for different methods tested on Gazebo data set, where bold numbers denote the best performance.}
\begin{tabular}{ccccccccc} % |c|c|c|c|c|c|c|c|c|
\toprule 
 & Initial & K-means &JRMPC &LRS& MATrICP&TMM & EMPRM & 3DMNDT\\
\midrule 
 $e_{\bf R} $ (rad.)&0.0241 &0.0097 &0.0216 &0.0092  & 0.0089 &0.0145 &0.0087 & \textbf{0.0082}\\
\midrule 
$e_t$ (m) &0.1069 &0.0318 &0.1005 & 0.0557 &   0.0511 &0.0630& \textbf{0.0185} & 0.0188\\ 
\midrule 
Time (min.) & / &0.8817 &7.8113 & 1.6324 & 1.9533 & 19.2284&  \textbf{0.4258} &0.6832  \\ 
\bottomrule
\end{tabular}
\label{Tab:SLAM}
\end{table*}

\subsubsection{Robustness to data noise} 
To illustrate its robustness to data noises, 3DMNDT as well as other registration methods are tested on object data sets added with random Gaussian noises. Considering randomness, each group of experiment is carried out by 30 independent tests. Experimental results are reported in the form of averaging registration errors. Tables \ref{Tab:50dB} demonstrate comparison results under two levels of noises. As illustrated in Tables \ref{Tab:50dB}, the accuracy of all registration methods decreases with the increase of noise level. In general, EMPMR and 3DMNDT are more robust to data noise than other registration methods.

Theoretically, JRMPC and TMM are probabilistic methods, where data noises are considered in the mixture model. However, they are more likely trapped into local minimum due to massive model parameters required to be optimized. Therefore, they seem to be sensitive to data noise. Compared with JRMPC and TMM, EMPMR requires to optimize fewer model parameters and it is more likely to obtain promising registration results. Since EMPRM directly formulates data noise in the objective function, it is very robust to data noise. Although both MATrICP and LRS do not take data noise into consideration, they seem robust to data noise for some data sets. This is because these two methods recover global motions from a set of relative motions and the influence of data noise on multi-view registration may be eliminated by the motion averaging or LRS matrix decomposition algorithm. However, high level noises may result in unreliable relative motions, which can lead to the failure of multi-view registration. Meanwhile, K-means also does not consider data noises, so it is less robust than EMPMR. 

Different from EMPRM, 3DMNDT does not directly formulate data noises in the objective function. However, it utilizes NDT, which allows data to contain noise. Therefore, it is also robust to data noise and is able to achieve promising registration results under different levels of Gaussian noises. 

 \subsection{Environment data set}   
 
Usually, multi-view registration methods can be potentially applied to SLAM. To illustrate its SLAM performance, 3DMNDT as well as other methods are tested on Gazebo data set. This data set is acquired by mobile robot equipped with the laser range finder in outdoor environment, where the robot mainly moved on the 2D plane ground and tracked the path to form a closed loop ($4 \times 5 \times 0.09$m). During robot movement, 31 range scans were acquired with ground truth of location, which were recorded by IMU and GPS. 

 \begin{figure*}[!t]
\centering
\subfigure[]{\includegraphics[scale=0.46]{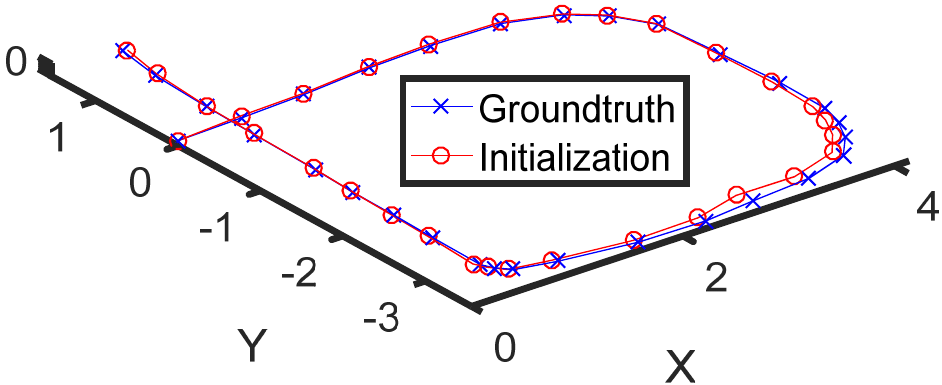}}
\subfigure[]{\includegraphics[scale=0.46]{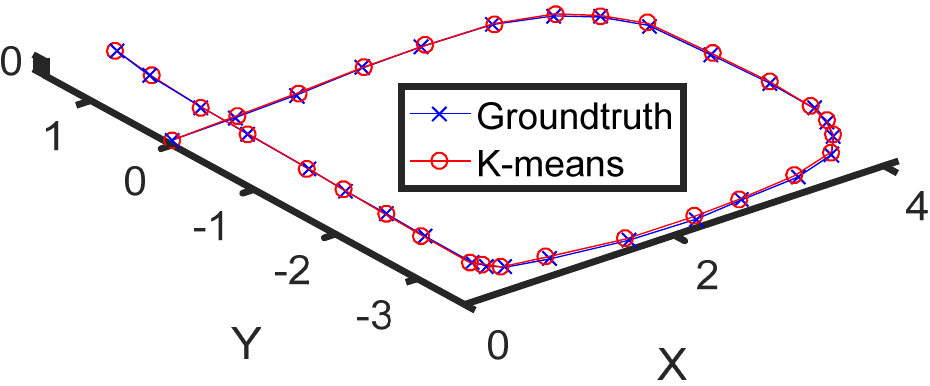}}
\subfigure[]{\includegraphics[scale=0.46]{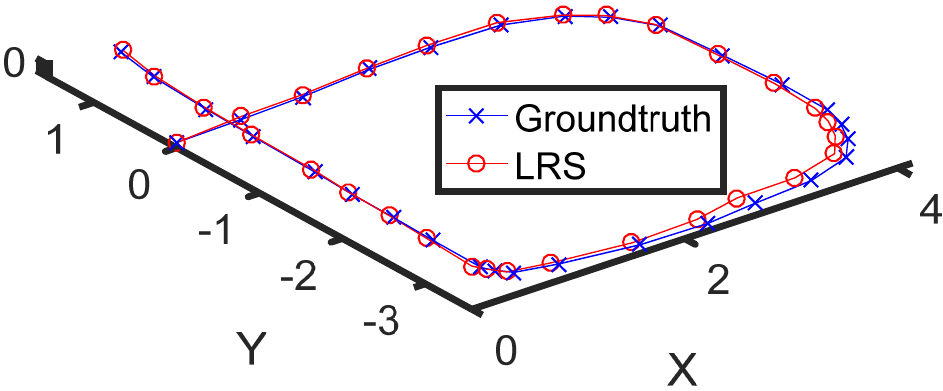}}
\subfigure[]{\includegraphics[scale=0.46]{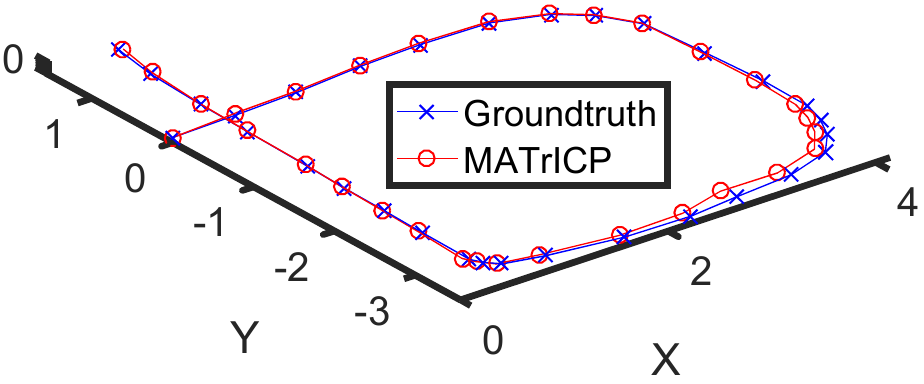}}
\subfigure[]{\includegraphics[scale=0.46]{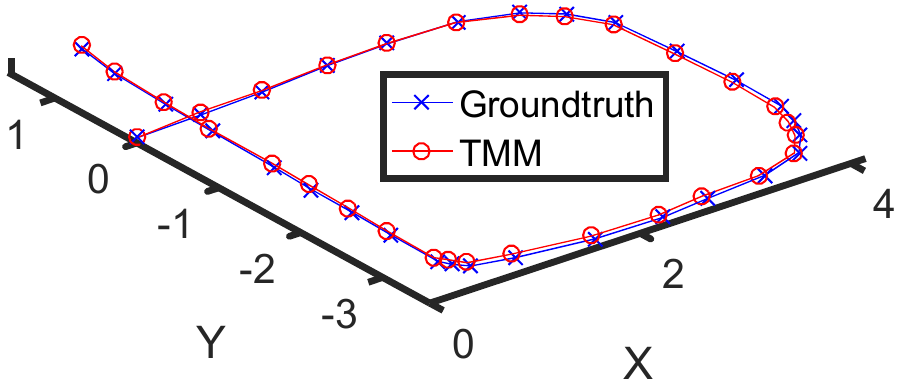}}
\subfigure[]{\includegraphics[scale=0.46]{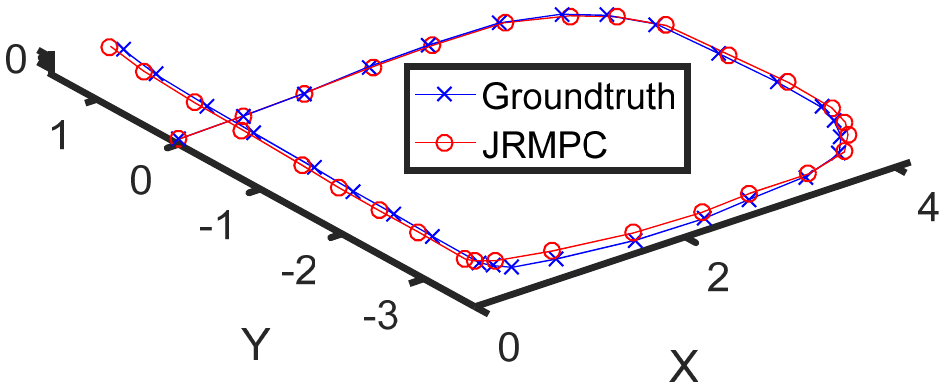}}
\subfigure[]{\includegraphics[scale=0.46]{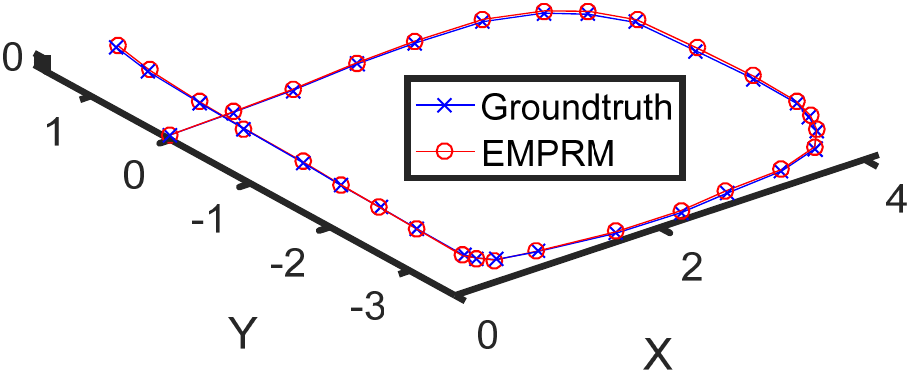}}
\subfigure[]{\includegraphics[scale=0.46]{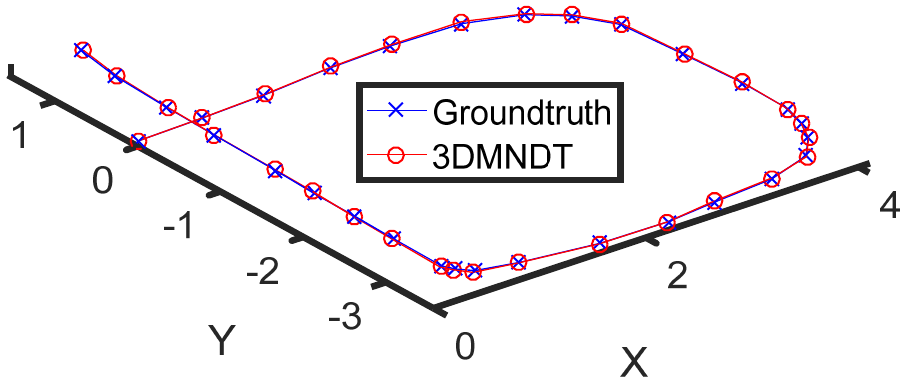}}
\caption{Illustration of localization results for different methods tested on Gazebo data set. (a) Initialization. (b) K-means result. (c) LRS result. (d) MATrICP result. (e) TMM result. (f) JRMPC result. (g) EMPRM result. (h) 3DMNDT result.}
\label{Fig:SLAM}
\end{figure*}

Before registration, data point indicated ground surface should be filtered by the height value and initial rigid transformations are provided by the TrICP algorithm. Table \ref{Tab:SLAM} illustrates the localization errors of all compared methods. For visual comparison, Fig. \ref{Fig:SLAM} displays corresponding localization results for these methods. As shown in Table \ref{Tab:SLAM} and Fig. \ref{Fig:SLAM}, 3DMNDT as well as EMPRM is more accurate and efficient than other compared methods. It is worth noting that both MATrICP and LRS only tend to reduce the localization error around the closed loop. This is because the closed loop introduces more available relative motions, which tend to substantially reduce the localization error around the closed loop. While, other methods are able to reduce localization error along the whole robot path. Among these methods, 3DMNDT and EMPRM can achieve SLAM with the best performance.
Therefore, 3DMNDT has the potential for the SLAM application.
    
\section{Conclusions}
Under the framework of NDT, this paper proposes a novel method for the multi-view registration. To the best of our knowledge, this is the ﬁrst time the NDT is applied to multi-view registration. Instead of space division, it utilizes k-means clustering algorithm to cluster data points and then computes NDT for each cluster. Compared with regular division, the K-means clustering can avoid unbalanced division of data points. Then, it formulates the multi-view registration problem by the NDT-based likelihood function, which is maximized by the Lie algebra solver. Different from the Newton's solver, the Lie algebra solver does not require to calculate Jacobian matrix and Hessian matrix, which is tedious and error-prone in 3D case. Experimental results tested on bench mark data sets illustrate that the proposed method can achieve state-of-the-art performance. What's more, it can be applied to SLAM. 

\section*{Acknowledgment}
This work is supported by the National Natural Science Foundation of China under Grant No. 61573273.
	
	% or
	
	

	\ifCLASSOPTIONcaptionsoff
	\newpage
	\fi

	
	
	\bibliographystyle{IEEEtran}
	\bibliography{IEEEfull}
	

	%\newpage

	
	%\vfill
	

\end{document}